\documentclass[preprint,12pt]{elsarticle}

\usepackage{amssymb}
\usepackage{amsmath}
\usepackage{booktabs}     
\usepackage{subcaption}   
\usepackage{multirow}      
\usepackage{tabularx} 
\usepackage{graphicx}    
\usepackage{float}  

\usepackage{algorithm}
\usepackage{algpseudocode}

\usepackage{pifont}  

\usepackage[hidelinks]{hyperref}

\journal{Expert Systems with Applications }

\begin{document}

\begin{frontmatter}

\title{An Efficient Deep Template Matching and In-Plane Pose Estimation Method via Template-Aware Dynamic Convolution}

\author[label1]{Ke Jia\fnref{equal1}}
\author[label1]{Ji Zhou\fnref{equal1}}
\author[label1]{Hanxin Li}
\author[label1]{Zhigan Zhou}
\author[label2]{Haojie Chu}
\author[label1]{Xiaojie Li\corref{cor1}}
\ead{lixj@cuit.edu.cn}

\fntext[equal]{These authors contributed equally to this work.}
\cortext[cor1]{Corresponding author.}

\affiliation[label1]{
          organization={School of Computer Science, Chengdu University of Information Technology},
          city={Chengdu},
          postcode={610225},
          state={Sichuan},
          country={China}
}

\affiliation[label2]{
          organization={Faculty of Mechanical Engineering and Mechanics, Ningbo University},
          city={Ningbo},
          postcode={315211},
           state={Zhejiang},
          country={China}
}

\begin{abstract}
In industrial inspection and component alignment tasks, template matching requires efficient estimation of a target's position and geometric state (rotation and scaling) under complex backgrounds to support precise downstream operations. Traditional methods rely on exhaustive enumeration of angles and scales, leading to low efficiency under compound transformations. Meanwhile, most deep learning-based approaches only estimate similarity scores without explicitly modeling geometric pose, making them inadequate for real-world deployment. To overcome these limitations, we propose a lightweight end-to-end framework that reformulates template matching as joint localization and geometric regression, outputting the center coordinates, rotation angle, and independent horizontal and vertical scales. A Template-Aware Dynamic Convolution Module (TDCM) dynamically injects template features at inference to guide generalizable matching. The compact network integrates depthwise separable convolutions and pixel shuffle for efficient matching. To enable geometric-annotation-free training, we introduce a rotation-shear-based augmentation strategy with structure-aware pseudo labels. A lightweight refinement module further improves angle and scale precision via local optimization. Experiments show our 3.07M model achieves high precision and $\sim$14\,ms inference under compound transformations. It also demonstrates strong robustness in small-template and multi-object scenarios, making it highly suitable for deployment in real-time industrial applications. The code is available at: \url{https://github.com/ZhouJ6610/PoseMatch-TDCM}.

\end{abstract}

\begin{graphicalabstract}
\begin{figure}[htbp]
\centering
\includegraphics[width=1\linewidth]{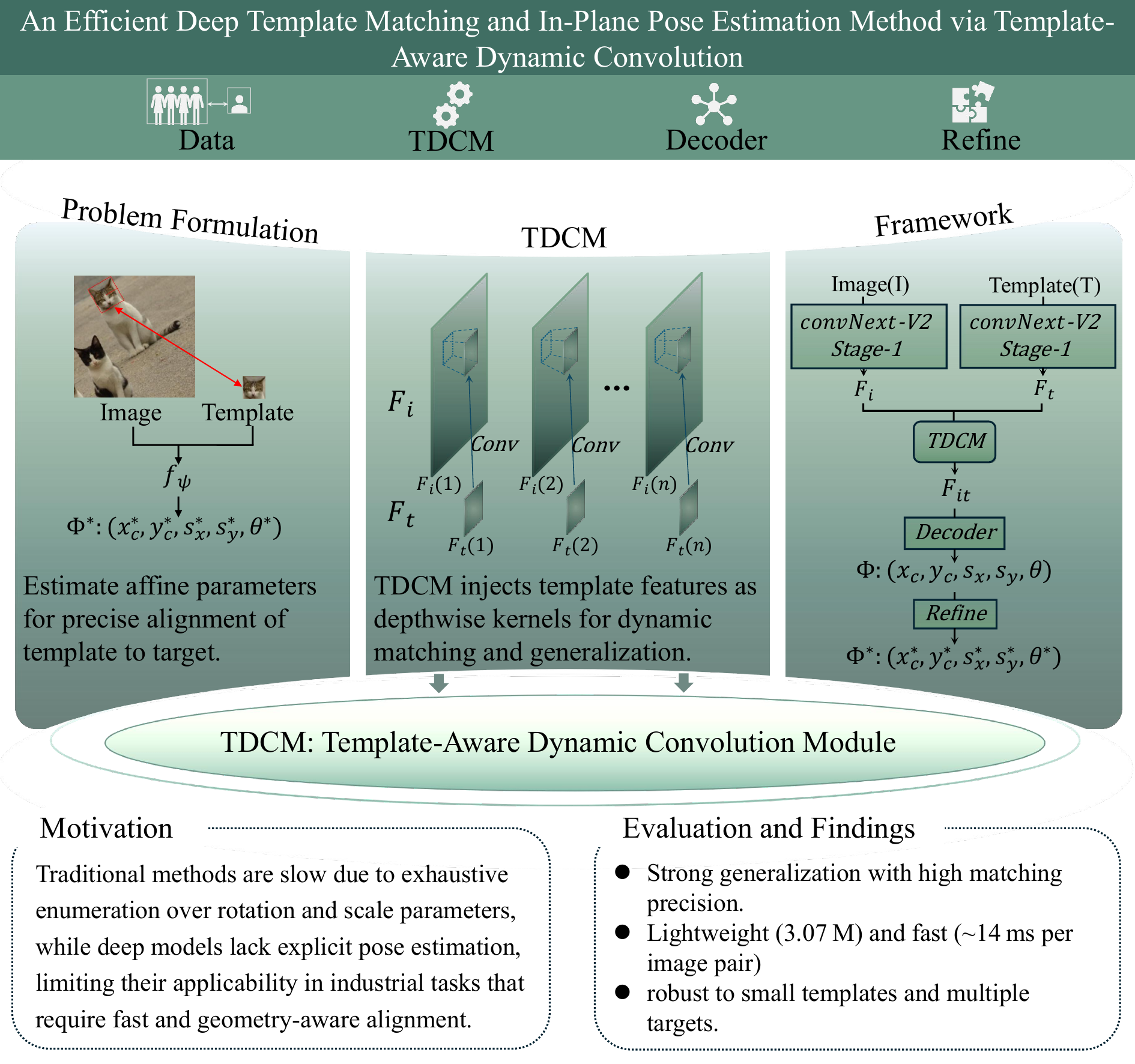}
\end{figure}

\end{graphicalabstract}
\begin{highlights}
\item End-to-end estimation of 2D geometric pose for planar template matching. 
\item TDCM enables strong generalization to unseen targets with efficient matching.
\item Compact 3.07M model achieves robust matching and real-time speed under transformations.
\item A refinement module improves angle-scale estimation via local geometric fitting.
\item Structure-aware pseudo labels enable self-supervised training without annotations.
\end{highlights}


\end{frontmatter}


\section{Introduction }
Template matching is widely used in industrial tasks such as quality inspection~\cite{Barone2020,Chen2025,BrakeYang2025}, precise component alignment~\cite{Le2020APCBAlignmentSystem}, and structural registration~\cite{Shahsavarani2024}, where the goal is to accurately identify the position, rotation, and scale of a target object under complex backgrounds to support precise downstream operations. This task not only requires accurate estimation of spatial location and geometric transformation but also demands strict control over inference efficiency to meet industrial deployment requirements. The core challenge lies in achieving fast matching with high precision. 

Traditional template matching methods based on local similarity measures—such as Sum of Squared Differences (SSD)~\cite{SSD2020,SSD-NCC} and Normalized Cross-Correlation(NCC)~\cite{NCC,SSD-NCC}—compute pixel-wise similarity between the template and the search image. While these approaches are simple and interpretable, they produce only response maps for localization and do not model geometric transformations explicitly. several methods~\cite{RLTM2007,le2022robot,ttm2024,FNCCR2024} go further by estimating rotation angles, but remain limited when scale variations are present. Some studies~\cite{FMT2005,GrayScale2007,TMbaseOrientGradient2012,CHEN2016207} further incorporate uniform scale modeling, yet they remain inadequate in real-world applications requiring independent scaling along different axes. Few existing works have addressed the complete estimation of 2D affine transformations. For example, Fast-Match~\cite{FAstMatch2013} adopts a branch-and-bound strategy combined with gradient-based descriptors to accelerate affine matching. Commercial software such as Halcon offers a Shape-Based Matching (SHM) module~\cite{halcon2024}, which integrates edge modeling with pyramid-based search to support transformation-aware detection. However, these methods rely on exhaustive parameter enumeration, leading to limited efficiency when handling compound transformations involving rotation, scaling, and translation.

In recent years, deep learning methods have achieved notable progress in template matching, particularly under non-rigid transformations, thanks to strong capabilities in feature extraction and similarity modeling. However, most existing approaches~\cite{qatm,SiameseSARwu2022,End-to-EndTMSiameseNetwork2022,CenterPointRemoteImagery2024,selfTM,SiameseNetworkYang2025},  focus on generating robust similarity response maps and do not explicitly predict geometric transformation parameters of the template, limiting their applicability in tasks requiring pose estimation. Therefore, within the task framework focused on fast and accurate template matching with pose estimation. To the best of our knowledge, existing deep learning methods have yet to establish an effective modeling mechanism. The lack of geometric state prediction remains a critical gap in current research. 

Keypoint-based methods such as SIFT~\cite{sift2004} and ORB~\cite{rublee2011orb}, as well as learned detectors like SuperPoint~\cite{superpoints2018} and R2D2~\cite{R2D22019}, provide a degree of invariance to scale and rotation through local feature matching.  However, these approaches often rely heavily on rich textures and local structures. Consequently, they struggle with small-sized templates due to insufficient texture information for reliable keypoint detection. Additionally, their reliance on global descriptor matching limits their ability to support multi-object scenarios.

To address these challenges, we propose a lightweight end-to-end template matching method that directly regresses the center position and geometric transformation parameters of the target, eliminating the need for exhaustive angle and scale enumeration. We propose TDCM to encodes template features as learnable convolutional kernels and integrates them into the search image using depthwise separable convolutions. Unlike methods that rely on fixed category embeddings or offline template encoding, our model adopts an explicit two-branch design, where the template is dynamically injected as a conditional input during inference. This design, inspired by prompt learning, guides the network to learn generalizable matching strategies rather than memorizing template appearances, thus enabling robust generalization to unseen templates. During training, we construct a self-supervised learning framework based on rotation-shear transformations and structure-aware pseudo-labels to guide the network in learning spatial correspondences between the template and the target without requiring manual annotations. A lightweight geometric refinement module is further introduced to optimize angle and scale predictions in a local neighborhood, enhancing final matching precision. Experimental results demonstrate that our lightweight 3.07M model surpasses Halcon’s Shape-Based Matching module in precision, while achieving significantly faster inference—only  $\sim$14\,ms per image pair. The method also maintains strong stability on small-sized templates and in multi-object matching scenarios. The main contributions of this work are summarized as follows:
\begin{itemize}
\item We propose a deep learning-based unified regression framework for template matching, which directly estimates five geometric parameters in an end-to-end manner, including the target center coordinates $(x_c, y_c)$, rotation angle $\theta$, and independent horizontal and vertical scales $(s_x, s_y)$.
\item We propose TDCM, which performs dynamic matching by injecting template features as depthwise kernels, enabling generalization to unseen targets.
\item We introduce a lightweight refinement module for local pose correction, and a self-supervised framework based on rotation-shear augmentation with structure-aware pseudo-labels to enable geometric-annotation-free training.
\end{itemize}

\section{Related Work}
\label{sec:related_work}
Template matching methods have been extensively studied across both traditional and deep learning domains. While some methods focus on robustness to non-rigid deformations such as bending or warping, many real-world applications demand accurate modeling of rigid geometric transformations like rotation and scaling. To better highlight how existing approaches address geometric transformations of increasing complexity, we categorize prior works into four levels: (1) basic localization methods that do not model transformations, (2) methods that explicitly estimate rotation, (3) methods that support rotation along with uniform scaling, and (4) methods capable of modeling both rotation and dual-axis scaling. 

\subsection{Localization without Transformation Modeling}

A large body of early template matching methods relied on local similarity measures to compute dense response maps between the template and the search image. Classic approaches such as SSD~\cite{SSD2020,SSD-NCC} and NCC~\cite{NCC,SSD-NCC} evaluated pixel-wise similarity directly, and many subsequent variants—e.g., BBS~\cite{BBS2015} and DDS~\cite{DDS2017}—aimed to improve robustness to illumination changes or background clutter. However, these methods did not explicitly model geometric transformations, and their matching was limited to the original template scale and orientation. To improve efficiency—particularly in high-resolution images—several acceleration strategies were proposed. Some works leveraged Fast Fourier Transform (FFT)~\cite{yoo2009fastCNN,kaso2018fft_ncc} to accelerate correlation computation in the frequency domain, effectively reducing the cost of sliding-window operations. Other approaches, such as approximate nearest neighbor search~\cite{diwu2018,vqnnf2023} and Segment-NCC~\cite{segmentNCC}, exploited local structure or feature redundancy to prune the search space and speed up template matching. While these techniques significantly reduced inference time, they still operated under rigid template assumptions and did not handle geometric variations such as rotation or scaling.

Recent deep learning-based approaches largely followed the same paradigm by treating template matching as a dense similarity regression task. Siamese-based networks~\cite{SiameseNetworkYang2025,End-to-EndTMSiameseNetwork2022}, Transformer-driven architectures~\cite{selfTM}, and quality-aware response modeling~\cite{qatm} all produced dense response maps without explicitly estimating pose parameters such as rotation or scale. Despite leveraging powerful feature representations, these models are fundamentally designed for position-sensitive matching under non-rigid appearance variations, rather than for estimating explicit geometric transformations. They lack mechanisms to directly model pose parameters such as rotation and scale. Although multi-scale or multi-angle enumeration can be applied as a workaround, such strategies are extremely inefficient. In rigid alignment scenarios, where both template localization and pose estimation are required, these similarity-based methods fall short due to their limited representational capacity. Some recent efforts, such as Gao et al.~\cite{deepTM}, specifically targeted cross-modality template matching by aligning features via coarse-to-fine affine transformations. However, such models were not suitable for standard mono-modal scenarios.

\subsection{Rotation Modeling}

To address the challenges of rigid matching under arbitrary orientations, several methods have been proposed to explicitly model rotation. Le and Lien~\cite{le2022robot} proposed a two-stage model that first located the target using deep feature-based template matching, then applied a self-rotation learning (SRL) network to regress the rotation angle. Tensorial Template Matching (TTM)~\cite{ttm2024} formulated template matching as a cross-correlation process over a rotation space, enabling direct estimation of the template's rotation angle. Annaby et al.~\cite{FNCCR2024} further proposed a Fast Normalized Cross-Correlation framework, which leveraged rotation-equivariant correlation in the frequency domain to achieve analytical robustness to rotation. These rotation-aware approaches offered better adaptability to orientation changes but were inherently constrained when applied to scenarios involving scale variations, particularly anisotropic (dual-axis) scaling.

\subsection{Rotation with Uniform Scaling}

Several methods extended rotation modeling by incorporating uniform scaling to improve robustness against combined transformations. The Fourier-Mellin Transform (FMT)~\cite{FMT2005} achieved rotation and scale invariance by converting the input image to log-polar coordinates in the frequency domain. Kim et al.~\cite{GrayScale2007} applied cascaded filters to exclude areas that had low probability of being selected as the final result. The method proposed by Konishi et al.~\cite{TMbaseOrientGradient2012} employed oriented gradient features with bitwise encoding, and achieved fast and precise matching by exhaustively comparing templates pre-rendered across multiple rotation angles and uniform scales. Chen et al.~\cite{CHEN2016207} introduced a fast CPU-based correlation alignment algorithm using image pyramids and SIMD optimization, and achieved real-time performance for rotation and single-axis scaling tasks. These methods represented a step forward compared to purely rotation-based approaches, but they typically assumed uniform or single-axis scaling. As a result, due to modeling only isotropic scaling, these methods remain inadequate when faced with full 2D rigid transformations involving translation, rotation, and anisotropic scaling

\subsection{Rotation with Dual-Axis Scaling}

A limited number of methods attempted to model more complex geometric transformations, including independent horizontal and vertical scaling, to enhance robustness in realistic scenarios. Halcon's commercial Shape-Based Matching (SHM) module~\cite{halcon2024} supported full rigid transformation modeling by combining edge-based template descriptions with a multi-level search over the parameter space. Despite its robustness, SHM relied heavily on exhaustive parameter enumeration, resulting in high computational cost. Korman et al.~\cite{FAstMatch2013} proposed Fast-Match, which accelerated affine template matching via randomized correspondence propagation and pruning strategies. Although the method significantly reduced the number of hypotheses compared to brute-force search, it still incurred substantial computation due to its reliance on iterative sampling and refinement. Notably, Fast-Match~\cite{FAstMatch2013} was primarily effective in low-texture or smooth regions, as it relied on the assumption of template smoothness to enable sparse yet accurate affine sampling. However, this assumption made the method unreliable under strong textures or high-frequency noise. While these methods provided high flexibility in modeling geometric transformations, their reliance on dense sampling or iterative refinement led to suboptimal inference efficiency—posing challenges for real-time prediction under compound transformations in industrial applications.

\section{Method}
\subsection{Problem Formulation }
This study focuses on the task of template matching and pose estimation, where given a template image $T \in \mathbb{R}^{H_T \times W_T \times C}$ and a search image $I \in \mathbb{R}^{H \times W \times C}$, the objective is to find a spatial transformation parameters $\Phi^{*}$ that aligns the template with the corresponding region in the search image. This transformation can be directly regressed by a matching model based on the input image pair:
\begin{equation}
\Phi^{*} = f_{\psi}(T, I)
\end{equation}
where $\Phi^{*}$ denotes the transformation parameters that align the template with its corresponding region in the search image. The function $f_{\psi}$ represents a parameterized template matching model with learnable parameters $\psi$. In the context of this paper, the spatial transformation $\Phi^{*}$ is parameterized as a rigid transformation tuple: the center position $(x^{*}_c, y^{*}_c)$, rotation angle $\theta^{*}$, and anisotropic scaling factors $s^{*}_x$ and $s^{*}_y$.

\subsection{Network Architecture}
\begin{figure}[H]
\centering
\includegraphics[width=1.0\linewidth]{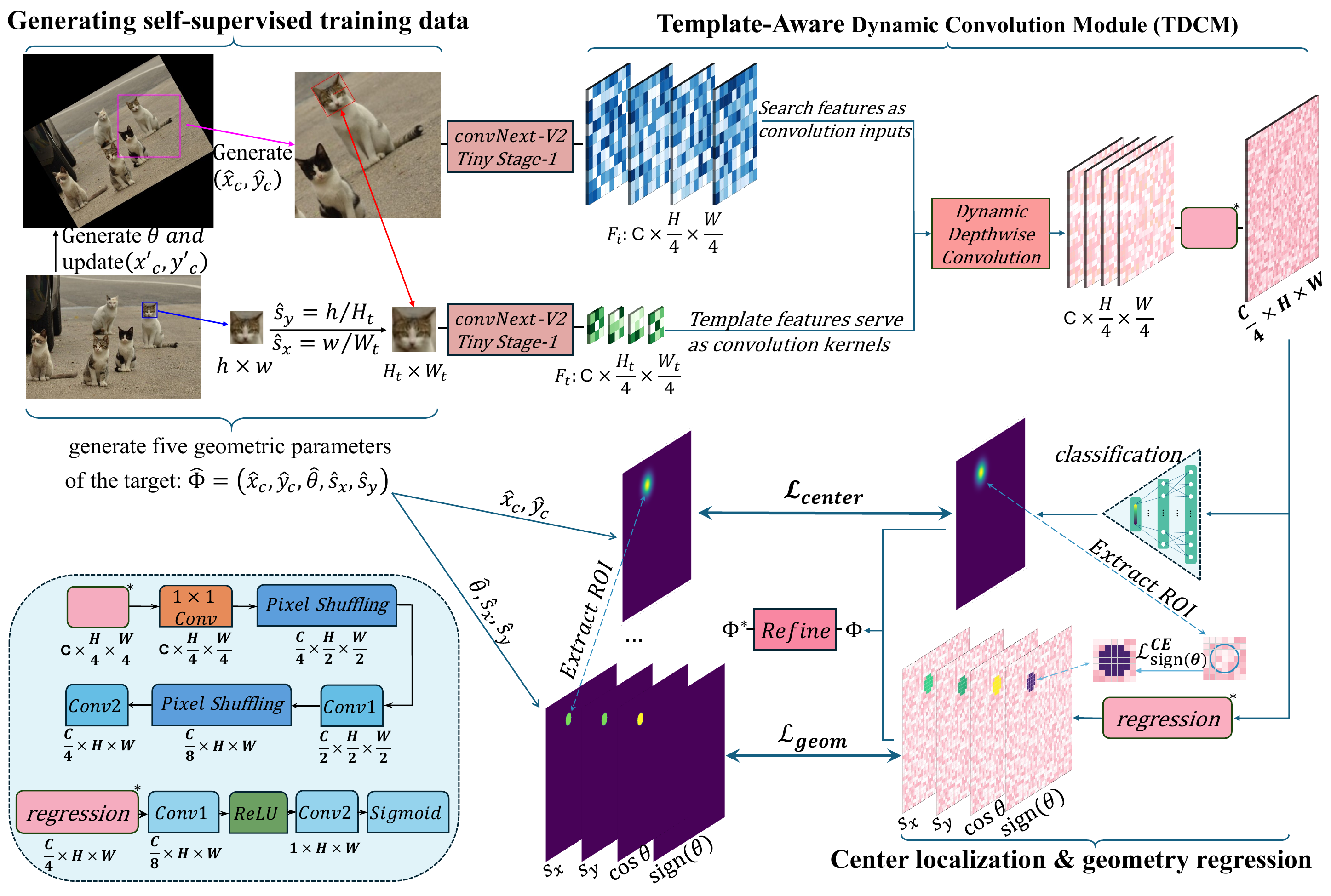}
\caption{Overview of the proposed framework. Shallow features are extracted from template and search images; template features are encoded as dynamic kernels and applied to search features, then decoded into response and parameter maps, followed by lightweight refinement for accurate pose estimation. }
\label{fig:framework}
\end{figure}

To enable efficient modeling of both the target location and geometric transformation parameters, we reformulate the task as a unified prediction framework of center localization and geometric parameter regression. Figure~\ref{fig:framework} illustrates the overall pipeline of the proposed method. This formulation draws inspiration from the regression paradigm commonly used in object detection, allowing the model to predict both the spatial position and transformation parameters of the target in a single forward pass, thus eliminating the need for exhaustive enumeration of rotation angles and scaling factors as in traditional methods. Moreover, shallow features—characterized by low semantic abstraction and high spatial fidelity—retain fine-grained structural cues from the original image, making them particularly suitable for position-sensitive template matching. Motivated by this, we exclusively utilize shallow features for the matching process, which not only preserves precise localization capability but also substantially reduces computational complexity.  

Building upon this foundation, we design a lightweight dual-branch network architecture composed of a template branch and a search image branch, both using the Stage-1 module of ConvNeXt-V2-Tiny~\cite{convNextV2_2023} for shallow feature extraction. To enable efficient inference and generalization to unseen templates, we propose the TDCM, which fully exploits template guidance by encoding its features as depthwise convolutional kernels applied to the shallow features of the search image. TDCM adopts depthwise separable convolutions as an essential component, which preserve structural alignment while further improving inference efficiency. The resulting structure-aware features are then passed to decoders that generate a confidence heatmap and spatial maps of geometric transformation parameters, enabling end-to-end pose-aware matching. Additionally, we incorporate a geometric refinement module that performs local optimization around the initially predicted angle and scale to further enhance matching accuracy.

 \subsection{Template-Aware Dynamic Convolution Module (TDCM) }
\begin{figure}[H]
\centering
\includegraphics[width=1\linewidth]{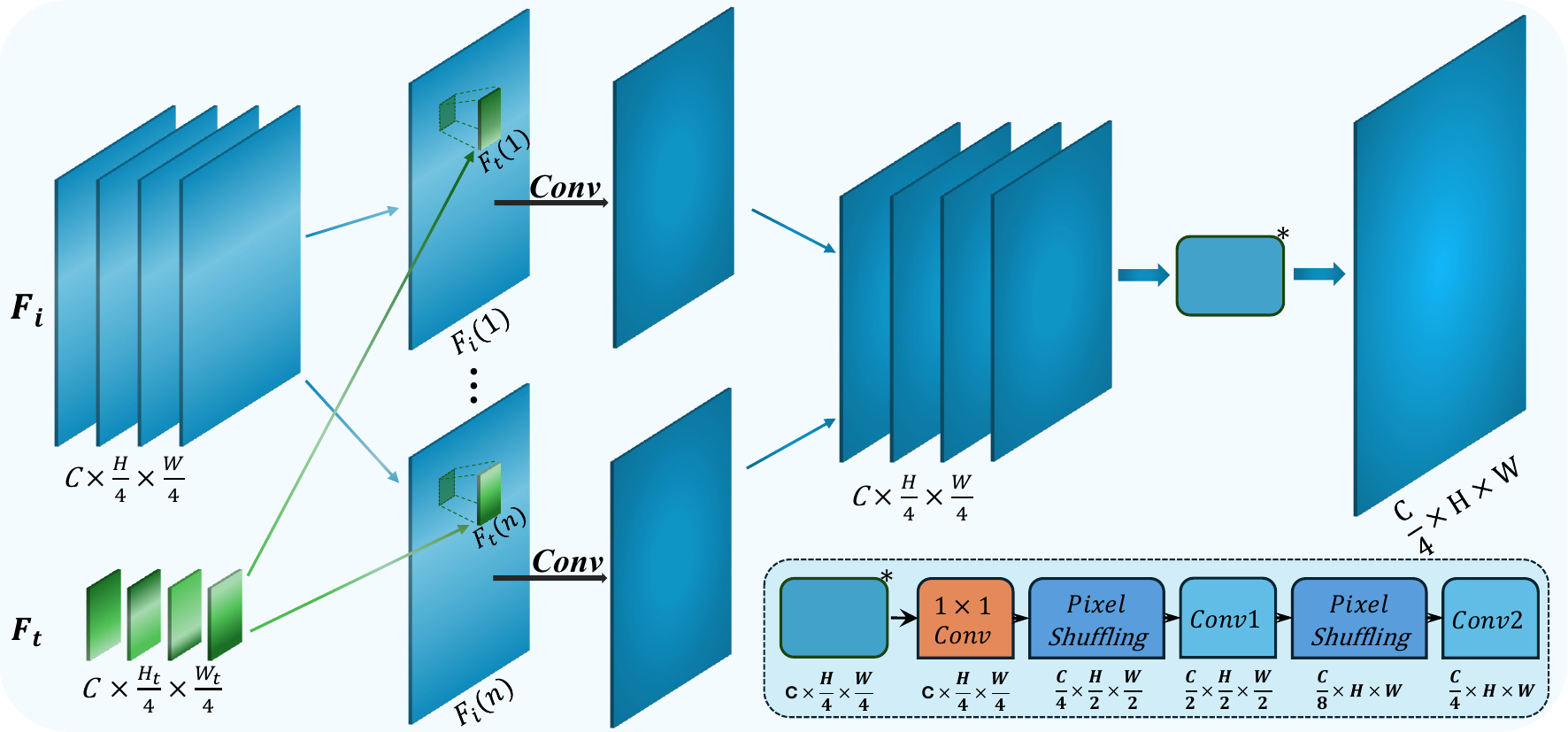}
\caption{Architecture of the Template-Aware Dynamic Convolution Module (TDCM). The template features is encoded as a dynamic convolution kernel and applied to the shallow search features via depthwise separable convolution. This enables structure-aligned feature fusion and facilitates pose-aware representation learning.}
\label{fig:tdcm}
\end{figure}

 \paragraph{TDCM Module Design}
In conventional deep learning-based template matching, models typically rely on fixed category labels or unified embedding spaces, which limits their adaptability to novel templates and hinders generalization in dynamic industrial scenarios where templates change frequently. To overcome this limitation and reduce dependency on memorized template features, we design a Template-Aware Dynamic Convolution Module (TDCM). This module dynamically encodes the input template features as depthwise convolutional kernels, which are then applied to the shallow features of the search image. By injecting template information at inference time, TDCM eliminates the need for static parameter modeling, enabling efficient structure-aware matching and generalization to unseen templates. It is important to note that in TDCM, the convolution kernels are directly derived from the template features, which are input-dependent and not trainable parameters, unlike conventional dynamic convolution methods. As illustrated in Figure~\ref{fig:tdcm}, the template image is first encoded into a feature tensor $F_t \in \mathbb{R}^{C \times \frac{H_t}{4} \times \frac{W_t}{4}}$, where $C$ is the channel dimension and $H_t, W_t$ are the template height and width. The search image is encoded into a tensor $F_i \in \mathbb{R}^{C \times \frac{H}{4} \times \frac{W}{4}}$. Each channel of the template feature map, denoted as $F_t(c)$, is used as a convolutional kernel to perform depthwise convolution over the corresponding channel of the search image feature map $F_i(c)$:
\begin{equation}
R(c) = F_t(c) * F_i(c)
\end{equation}
Where $R(c)$ denotes the channel-wise response map that captures the spatial correlation between the template and search features in channel $c$. These response maps are subsequently fused through a pointwise (1$\times$1) convolution and upsampled via a pixel shuffle operation to restore spatial resolution. TDCM yields a high-resolution spatial feature map that preserves positional information necessary for accurate localization and regression. By using dynamic depthwise separable convolution, The overall architecture of the TDCM module enables rapid response to rotated and scaled targets without relying on multi-scale or multi-angle enumeration, serving as a core component for enabling both efficient inference and generalization to arbitrary templates.

\paragraph{Working Principle of TDCM}

In the proposed TDCM module, convolution is employed as an efficient surrogate for similarity computation. The kernels are directly derived from the template features, so the resulting responses can be naturally interpreted as the semantic similarity between the template and the search features. 

From a mathematical perspective, this similarity aligns closely with the well-known normalized cross-correlation (NCC) used in classical template matching, such as \texttt{CV\_TM\_CCORR\_NORMED} in OpenCV. NCC can be formally expressed as
\[
R(x, y) = 
\frac{
\sum_{x', y'} \big( T(x', y') \cdot I(x+x', y+y') \big)
}{
\sqrt{
\sum_{x', y'} T(x', y')^2 \cdot 
\sum_{x', y'} I(x+x', y+y')^2
}
},
\]
where $T(x',y')$ denotes the template features and $I(x+x',y+y')$ denotes the local region of the search image at location $(x,y)$. The numerator corresponds to the dot product between the template and the search patch (which is exactly what convolution computes), while the denominator normalizes by their magnitudes. The numerator corresponds to the dot product between the template and the search patch (which is exactly what convolution computes), while the denominator normalizes by their magnitudes. This normalization makes the operation equivalent to cosine similarity, since the vector inner product can be written as
\[
\langle w, x \rangle = \|w\| \cdot \|x\| \cdot \cos \theta,
\]
where the cosine term measures the similarity of $w$ and $x$ vectors. In other words, NCC can be understood as a variant of cosine similarity, which explains why it provides a meaningful measure of feature similarity in template matching. Such dot-product-based similarity is also widely adopted in attention mechanisms. Moreover, convolution itself is equivalent to a sliding-window process, computing local similarity across spatial positions. When the normalization term is omitted, as in the unnormalized variants (e.g., \texttt{CV\_TM\_CCORR} or \texttt{CV\_TM\_CCOEFF}), only the dot product remains, further confirming the feasibility of using pure feature inner products for matching. 

Beyond similarity, convolution as a fundamental neural operator can capture richer feature interactions in deep representation space, offering more expressive and adaptive modeling capacity compared with handcrafted similarity metrics. Specifically, TDCM adopts a depthwise–pointwise design: depthwise convolution achieves channel-wise semantic alignment, while pointwise convolution enables efficient global fusion. This design both preserves the fidelity of template semantics and reduces computational overhead, making it suitable for real-time inference. Therefore, in our framework, convolution not only replaces traditional similarity computation but also provides an efficient and robust mechanism for template-aware matching.

\paragraph{TDCM vs. Existing Dynamic Convolutions}

Dynamic convolution has been extensively studied, with representative frameworks such as CondConv~\cite{CondConv2019}, DynamicFilterNet~\cite{DynamicFilterNet}, and DCNv4~\cite{DCNv4}. In these approaches, the “dynamic” behavior is realized indirectly, either by predicting proxy kernels from input features (CondConv, DynamicFilterNet) or by adjusting the sampling positions of fixed kernels (DCN series). In all such cases, the convolution kernel itself is not inherently tied to the target semantics.

In contrast, the essence of TDCM lies in a different form of dynamic: the convolution weights themselves act as filters whose response patterns adapt on demand according to the template. Rather than approximating a kernel via an auxiliary predictor, TDCM directly instantiates the template feature map as the convolution kernel and applies it to the search feature map. This ensures that the filter is inherently target-aware and preserves semantic information of the template throughout the matching process, rather than reducing it to potentially lossy surrogate parameters.

This fundamental difference yields three unique characteristics of TDCM: (1) it is explicitly designed for template matching rather than generic single-input recognition tasks, (2) it inherently requires dual inputs (template and search) to establish explicit target-conditioned interaction, and (3) it adopts a depthwise–pointwise design for efficient per-channel matching and real-time inference. Table~\ref{tab:dynamic_conv_comparison} summarizes these distinctions, with the “Kernel Source” column highlighting whether kernels are directly inherited from template features (TDCM) or indirectly predicted from input features (other frameworks), and the “Task-Specific” column indicating whether the operator is explicitly optimized for template matching.
\begin{table}[H]
\centering
\caption{Comparison between existing dynamic convolution frameworks and TDCM.}
\label{tab:dynamic_conv_comparison}
\renewcommand{\arraystretch}{1.2} 
\resizebox{\textwidth}{!}{
\begin{tabular}{l|cccc}
\hline
\textbf{Method} & \textbf{Kernel Source} & \textbf{Dual-Input} & \textbf{Depthwise} & \textbf{Task-Specific} \\
\hline
CondConv\cite{CondConv2019}             & Predicted kernels & \ding{55} & \ding{55} & \ding{55} \\
DCNv4\cite{DCNv4}                & Predicted kernels & \ding{55} & \ding{55} & \ding{55} \\
DynamicFilterNet\cite{DynamicFilterNet}     & Predicted kernels & \ding{55} & \ding{55} & \ding{55} \\
TDCM (Ours)          & Template features & \ding{51} & \ding{51} & \ding{51} \\
\hline
\end{tabular}
}
\end{table}

\subsection{Training Data and Self-Supervised Pseudo Labeling }
 
\paragraph{Training Data}

Inspired by prior studies\cite{simple2021cvpr,spatialpooling2022pr,fnnet2025pr} in pseudo-labeling and adaptive augmentation strategies, we further develop an automated training strategy based on synthetic transformations to avoid the cost and impracticality of collecting manual annotations for geometric transformations. This strategy simulates diverse deformation scenarios and produces structurally aligned pseudo-labels, guiding the network to learn transformation-aware representations. Training pairs are constructed from the MS-COCO dataset by cropping a template from the bounding box, which is then resized to satisfy the depthwise convolution alignment constraint of $8n + 4$. The corresponding scale factors are computed accordingly. A random rotation $\hat\theta \in (-180^\circ, 180^\circ)$ is applied to both the image and the bounding box, after which a search region is cropped to fully include the transformed object while excluding any padded border artifacts. The center coordinate is recalculated in the transformed space. The detailed procedure for constructing training image pairs is summarized in Algorithm\textbf{~\ref{alg:pair-construction}}.

\begin{algorithm}[ht]
\caption{Construction of Training Image Pairs from MS-COCO}
\label{alg:pair-construction}
\begin{algorithmic}[1]
\Require Image $S$ and bounding box $B = (x, y, w, h)$
\Ensure Template $T$, Search image $S$, target parameters $\hat{x}_c, \hat{y}_c$, $\hat{s}_x$, $\hat{s}_y$, $\hat{\theta}$

\State Crop raw template $T_{\text{raw}}$ from $S$ using bounding box $B$
\State Resize $T_{\text{raw}}$ to $T \in \mathbb{R}^{H_t \times W_t}$, where $(W_t, H_t)$ satisfy $8n + 4$
\State Compute $\hat{s}_x \gets \frac{w}{W_t}$, $\hat{s}_y \gets \frac{h}{H_t}$
\State Sample $\hat{\theta} \sim \mathcal{U}(-180^\circ, 180^\circ)$
\State Rotate $S$ by $\hat{\theta}$ to obtain $S_{\hat{\theta}}$, in which $B$ becomes $B_{\hat{\theta}}$
\State Crop region $I$ from $S_{\hat{\theta}}$ that fully contains $B_{\hat{\theta}}$ and avoids blank borders
\State Compute center $(\hat{x}_c, \hat{y}_c)$ of $B_{\hat{\theta}}$ in $I$
\State \Return $T$, $I$, $\hat{x}_c, \hat{y}_c$, $\hat{s}_x$, $\hat{s}_y$, $\hat{\theta}$
\end{algorithmic}
\end{algorithm}

It should be emphasized that, although paste-based synthesis provides clearer visualization of matching results as shown in Figure ~\ref{fig:visExamples}, it introduces artificial boundaries and disrupts texture continuity, which leads to a domain gap when used for training (a phenomenon that we empirically validate in our experiments). Therefore, we do not adopt paste-based pairs for training. Instead, we apply affine transformations and crop rotated regions directly from the original image, enabling more realistic geometry and background consistency. 

\paragraph{Geometry-Aware Gaussian Heatmap}
\label{sec:label_generation}
To provide spatially-aware supervision for center localization, we employ a Gaussian heatmap whose response approximates the degree of overlap with the template. To incorporate geometric attributes, we deform a base isotropic Gaussian using an affine transformation matrix constructed from ground-truth pose parameters. A base Gaussian centered at $(x_c, y_c)$ is defined as:
\begin{equation}
\mathcal{G}(x, y) = \exp\left(-\frac{(x - \hat{x}_c)^2 + (y - \hat{y}_c)^2}{2\sigma^2}\right)
\end{equation}
where $\sigma$ controls the spatial spread and $(\hat{x}_c, \hat{y}_c)$ denotes the target center. To encode orientation and anisotropic scaling, we construct an affine matrix:
\begin{equation}
\mathbf{A} = 
\begin{bmatrix}
\hat{s}_x \cos\hat\theta & -\hat{s}_y \sin\hat\theta \\
\hat{s}_x \sin\hat\theta & \hat{s}_y \cos\hat\theta
\end{bmatrix}
\end{equation}
where $\hat{s}_x$, $\hat{s}_y$, and $\hat\theta$ are the ground-truth scaling factors and rotation angle. This matrix is applied to deform the coordinate grid, producing an elliptical Gaussian:
\begin{equation}
\tilde{\mathcal{G}}(x, y) = \mathcal{G}(\mathbf{A}^{-1}(x - \hat{x}_c, y - \hat{y}_c))
\end{equation}
where $\mathbf{A}^{-1}$ warps each location into the canonical Gaussian coordinate space. The resulting heatmap encodes both positional and geometric cues, providing fine-grained supervision for robust spatial representation learning. During training, pixels with response scores above a threshold (e.g., 0.5) are selected as valid samples.

While the structure-aware heatmap guides attention to object-centric regions, it lacks the capacity to represent fine-grained geometric transformations due to its limited resolution and soft responses. To address this limitation, we assign dense regression targets to all ground-truth object center locations with activation values greater than 0.5. At these valid locations, the pseudo-label maps are filled with the normalized values of the geometric transformation parameters, including $s_x$, $s_y$, $\mathrm{sign}(\theta)$, and $\cos(\theta)$. This complementary supervision enables accurate learning of geometric transformations beyond the heatmap's expressiveness and use of dense spatial parameter maps allows for effective handling of multi-object scenarios.

\subsection{Loss Functions }

To enhance robustness in center localization, we propose a soft-label classification loss inspired by CenterNet's heatmap supervision. It replaces hard one-hot labels with soft masks to capture spatial uncertainty, and applies asymmetric penalties to positive and negative pixels, normalized by the number of positives, making it well-suited for sparse foreground detection under class imbalance. The loss is defined as:
\begin{equation}
\mathcal{L}_{\text{center}} = -\frac{1}{N_{\text{pos}}} \sum_{x,y} \left[
\mathbf{1}_{Y_{x,y} \ge 0.5} \cdot \mathrm{BCE}(\hat{Y}_{x,y}, Y_{x,y}) +
\mathbf{1}_{Y_{x,y} < 0.5} \cdot \log(1 - \hat{Y}_{x,y})
\right]
\end{equation}
where \( \hat{Y}_{x,y} \) is the predicted heatmap, \( Y_{x,y} \) is the soft ground truth, and \( N_{\text{pos}} \) denotes the number of foreground pixels with \( Y_{x,y} \ge 0.5 \). For geometric supervision, the network predicts four additional parameter maps: the cosine of the rotation angle $\cos\theta$, The sign of the rotation angle $\text{sign}(\theta)$, and the scaling factors ${s}_x$ and ${s}_y$. These values are only supervised within the valid region defined by the heatmap threshold.

The rotation sign is trained using a cross-entropy loss, while the other parameters are optimized using $\ell_1$ regression losses. The total geometric loss is given by:

\begin{equation}
\mathcal{L}_{\text{geom}} = \mathcal{L}_{\cos\theta}^{\ell_1} + \mathcal{L}_{\text{sign}(\theta)}^{\text{CE}} + \mathcal{L}_{{s}_x}^{\ell_1} + \mathcal{L}_{{s}_y}^{\ell_1}
\label{eq:geom_loss}
\end{equation}

Finally, the overall loss combines all components with equal weights:

\begin{equation}
\mathcal{L}_{\text{total}} = \mathcal{L}_{\text{center}} + \mathcal{L}_{\text{geom}}
\label{eq:total_loss}
\end{equation}

To clarify, we assign equal weights to all five optimization terms, including the center localization loss and the four geometric regression losses, as defined in Eq.~\ref{eq:geom_loss} and Eq.~\ref{eq:total_loss}. We also experimented with manually adjusting the relative weights but observed negligible performance gains. Therefore, we adopt equal weighting throughout the experiments to maintain simplicity and training stability.

\subsection{Geometric Refinement Module}

Although the backbone can regress the approximate target position and geometry, prediction errors often arise under strong transformations or weak textures. This stems from the lack of inherent rotation and scale invariance in convolutional layers, making the regression essentially a fitting process influenced by limited receptive fields. In contrast, traditional methods mitigate such errors by explicitly enumerating transformation parameters to align the template via affine warping before computing similarity, thereby compensating for large appearance variations. Inspired by this, we introduce a lightweight geometric refinement module that locally aligns the predicted region and re-evaluates similarity to correct residual errors, especially in angle and scale. 

Specifically, given the preliminary outputs \((x_c, y_c)\), \(s_x\), \(s_y\), and \(\theta\) from the regression head, we perform a local exhaustive search to optimize the rotation angle. To this end, a candidate angle set \(\Theta = \{\theta_i\}\) is constructed by uniformly sampling within a small neighborhood (e.g., \(\pm 20^\circ\) with 1° steps) centered around the initial estimate. For each candidate angle \(\theta_i\), the template is rotated accordingly, resized based on the predicted scale parameters \(s_x, s_y\), and then pasted at the predicted center position \((x_c, y_c)\) to generate a new candidate template \(T_i\). A rectangular region \(\Omega_i = \{(x, y) \mid T_i(x, y) > 0\}\) is then extracted, and the similarity between the search image and each candidate template is computed using a cosine similarity metric guided by the transformed template mask, ensuring feature alignment within the expected region. The angle corresponding to the highest similarity is selected as the final output. This process is formally defined as:

\begin{equation}
\theta^* = \arg \max_{\theta_i \in \Theta} \mathrm{sim}(I, T_i, \Omega_i)
\label{eq:refine_theta}
\end{equation}

where \(\mathrm{sim}(\cdot)\) denotes the cosine similarity over the masked region \(\Omega\), defined as:
\begin{equation}
\mathrm{sim}(A, B, \Omega) = \frac{\langle A_\Omega, B_\Omega \rangle}{\|A_\Omega\| \cdot \|B_\Omega\|}
\label{eq:refine_sim}
\end{equation}

where \(A_\Omega\) and \(B_\Omega\) represent the valid pixel sets of image \(A\) and \(B\) restricted to region \(\Omega\). If all candidate similarities fall below a preset threshold (e.g., 0.9), the initial prediction is retained to ensure stability. The refinement module is lightweight, hyperparameter-free, and performs local re-alignment by sampling a few candidate transformations around the initial estimate, thereby improving the overall pose estimation accuracy. The complete process is outlined in Algorithm\textbf{~\ref{alg:refine}}.

\begin{algorithm}[H]
\caption{Geometric Refinement via Angle Search}
\label{alg:refine}
\begin{algorithmic}[1]
\Require Search image $\mathcal{I}$, template $\mathcal{T}$, initial estimate $(x_c, y_c, \theta, s_x, s_y)$
\Ensure Refined rotation angle $\theta^*$

\State Set step size $\Delta \theta$ and search radius $r$
\State Generate angle set $\Theta = \{\theta_i \mid \theta_i \in [\theta - r, \theta + r], \theta_i = \theta + k \cdot \Delta \theta\}$

\ForAll{$\theta_i \in \Theta$}
    \State Apply affine transform: $\mathcal{T}_i \gets \text{Affine}(\mathcal{T}, x_c, y_c, \theta_i, s_x, s_y)$
    \State Extract mask region $\Omega_i$ corresponding to $\mathcal{T}_i$
    \State Compute masked cosine similarity: $S_i = \text{Sim}(\mathcal{I}, \mathcal{T}_i, \Omega_i)$
\EndFor

\State $\theta^* \gets \arg\max_{\theta_i \in \Theta} S_i$

\State \Return $\theta^*$
\end{algorithmic}
\end{algorithm}

\section{Experiments } 

Most deep learning-based methods focus on robustness under non-rigid transformations but lack explicit geometric modeling, making them unsuitable for tasks requiring precise pose estimation. Moreover, these methods usually rely on exhaustive angle and scale enumeration to handle the geometric variations considered in this work, which leads to significantly lower efficiency and does not align with the lightweight and real-time objectives of our approach. Therefore, we do not include such methods in our comparison and instead compare with baselines explicitly handling rigid transformations.

This section presents a comprehensive evaluation of the proposed method in terms of both computational efficiency and matching precision. As discussed in Section~\ref{sec:related_work}, most deep learning-based methods focus on robustness under non-rigid transformations but lack explicit geometric modeling, making them unsuitable for tasks requiring precise pose estimation. Moreover, these methods must rely on exhaustive angle and scale enumeration to handle the geometric variations considered in this work, which leads to significantly lower efficiency and does not align with the lightweight and real-time objectives of our approach. Therefore, we do not include such methods in our comparison and instead compare with baselines explicitly handling rigid transformations. Specifically, we use an enhanced version of NCC as a classical baseline, where exhaustive search over rotation and scaling parameters is performed using OpenCV’s normalized cross-correlation implementation\footnote{Implemented via \texttt{cv2.matchTemplate} with \texttt{cv2.TM\_CCOEFF\_NORMED}}. We also include Fast-Match~\cite{FAstMatch2013}, a representative academic method, and Halcon’s Shape-Based Matching (SHM) module as a commercial solution. The experiments are organized as follows:

\begin{itemize}
    \item Performance under different levels of geometric transformations;
    \item Multi-instance matching evaluation;
    \item Adaptability across different template sizes.
\end{itemize}

\begin{figure}[H]
\centering
\includegraphics[width=1\linewidth]{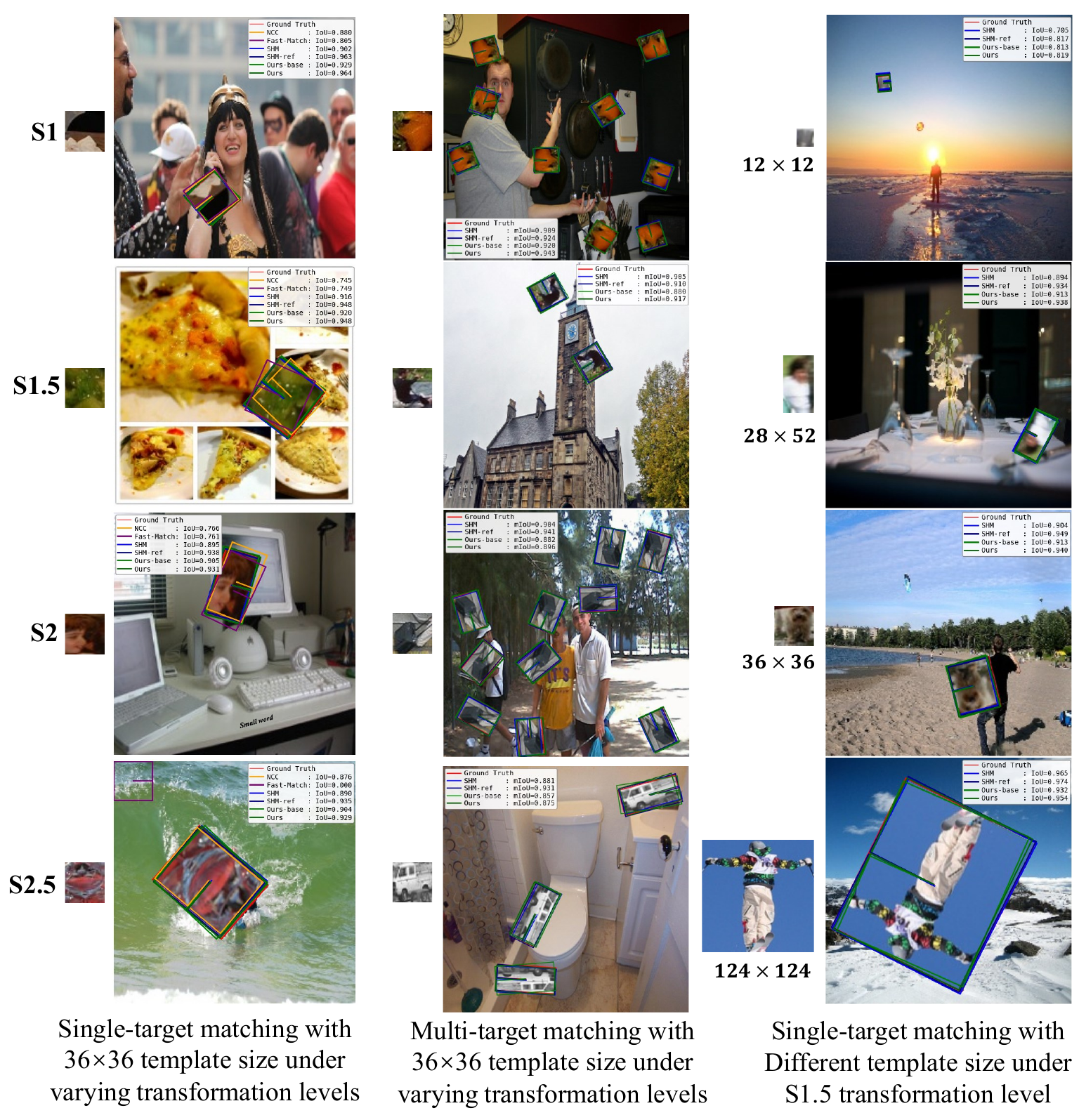}
\caption{Representative matching results under various transformation settings.}
\label{fig:visExamples}
\end{figure}

Experimental results show that our method consistently achieves strong precision and stable matching speed under different transformation conditions. In particular, it significantly reduces inference time in cases involving wide-range rotation and scaling, demonstrating favorable computational efficiency and deployment potential. Fig.~\ref{fig:visExamples} shows representative matching results across different transformation scenarios, generated with paste-based synthetic data. 

\subsection{Experimental Setup}

All experiments are implemented using the PyTorch framework and conducted on a workstation equipped with an Intel Core i9-14900KF CPU (24 cores, 32 threads). Notably, all inference is performed exclusively on the CPU without GPU acceleration. The backbone network is initialized with pretrained weights from the Stage-1 module of ConvNeXt-V2-Tiny~\cite{convNextV2_2023} , and only shallow features are extracted to maintain model lightweightness. The total number of parameters is 3.07M.

Both training and validation samples are automatically generated from the MS-COCO images using our proposed procedure described in Section~\ref{sec:label_generation}, covering full $360^{\circ}$ rotations and multiple scaling intervals. For fair comparison, both NCC, Fast-Match and SHM are evaluated on the same dataset and carefully tuned to achieve their optimal performance.  

In addition to standard template sizes $36\times36$, we evaluate the performance under various template configurations, including sizes of $12\times12$, $28\times52$, and $124\times124$. To comprehensively assess the proposed method’s robustness and efficiency under geometric transformations, we define four levels of increasing transformation complexity:

\begin{itemize}
    \item \textbf{S1}: rotation only (scale fixed at 1)
    \item \textbf{S1.5}: rotation with mild scaling ($0.8 \sim 1.5$)
    \item \textbf{S2}: rotation with moderate scaling ($0.5 \sim 2$)
    \item \textbf{S2.5}: rotation with large scaling ($0.4 \sim 2.5$)

\end{itemize}

\subsection{Matching Performance under Varying Transformation Levels}

The proposed method demonstrates strong accuracy and significant inference efficiency across all transformation conditions. As shown in Table \ref{tab:transformation_levels}, under all compound transformation, our method consistently demonstrates significant efficiency gains. For instance, under the most challenging setting (S2.5), our method achieves a more than 180$\times$ speed-up over SHM, demonstrating its robustness to wide-range angle-scale transformations. This speed advantage stems from the fundamental difference in geometric modeling strategies. SHM relies on multi-resolution search and exhaustive enumeration of transformation parameters. As the transformation range increases, SHM internally increases the density of angle and scale sampling to preserve accuracy, which causes inference time to grow exponentially. In contrast, our method directly regresses geometric parameters in a single forward pass, using a lightweight refinement module only for local pose optimization. This architecture ensures that computational complexity remains nearly constant, regardless of transformation intensity. Both NCC and Fast-Match also exhibit slow inference caused by exhaustive enumeration or iterative optimization.

In terms of matching accuracy, our method consistently outperforms SHM, NCC, and Fast-Match, achieving higher mean Intersection-over-Union (mIoU) across all evaluation settings. The relatively low accuracy of Fast-Match may be attributed to the strong texture patterns in our test data, which deviate from the low-texture assumption under which it was originally designed. Further analysis reveals that while SHM performs better in rotation and scale estimation, our approach delivers more precise center localization, complementing its geometry modeling with stronger spatial response accuracy. Moreover, as shown in Table~\ref{tab:transformation_levels}, increasing the scaling range from S1 to S2.5 introduces stronger geometric variations during augmentation, which leads to a noticeable decrease in the matching accuracy of TDCM. As expected, both the traditional NCC and the classical Fast-Match method perform poorly across all scenarios. 

\begin{table}[H]
\centering
\caption{Matching performance under varying transformation levels (S1–S2.5).}
\label{tab:transformation_levels}
\renewcommand{\arraystretch}{1.2}
\resizebox{\textwidth}{!}{
\begin{tabular}{c l c c c c c c}
\toprule
\textbf{Setting} & \textbf{Method} & \textbf{Loc.Err ↓} & \textbf{ScaleErr ↓} & \textbf{Rot.Err ↓}& \textbf{mIoU ↑} & \textbf{Time(ms) ↓}& \textbf{Succ.(\%) ↑}\\
\midrule
S1     & NCC  & 2.05 & --   & 4.73°& 0.855 & 422& 73.1\\
 & Fast-Match  & 3.52& --& 7.56°& 0.797& 232&74.2\\
       & SHM  & 1.46 & --   & \textbf{0.24°}& 0.906 & \textbf{4.49}& 98.9\\
       & Ours & \textbf{0.44} & -- & 1.03°& \textbf{0.955}& 11.3& \textbf{99.6}\\
\midrule
S1.5   & NCC  & 6.75 & 0.127 & 5.03°& 0.726 & 6996& 60.2\\
       & Fast-Match  & 4.13& 0.142& 9.03°& 0.756& 576& 84.7\\
       & SHM  & 1.50 & \textbf{0.027} & \textbf{0.47°}& 0.908 & 97.8& 97.3\\
       & Ours & \textbf{0.64} & 0.038 & 1.93°& \textbf{0.926}& \textbf{13.8}& \textbf{98.6}\\
\midrule
S2     & NCC  & 9.92 & 0.124 & 3.16°& 0.632 & 34932& 52.8\\
 & Fast-Match  & 4.56& 0.172& 9.88°& 0.677& 1886&62.7\\
       & SHM  & 1.74 & \textbf{0.034} & \textbf{0.48°}& 0.898 & 824& 98.1\\
       & Ours & \textbf{0.82}& 0.043& 2.80°& \textbf{0.900}& \textbf{14.2}& \textbf{99.0}\\
\midrule
S2.5   & NCC  & 11.22 & 0.122 & 3.54°& 0.616 & 73747& 43.9\\
 & Fast-Match  & 5.16& 0.198& 9.52°& 0.664& 3369&59.5\\
       & SHM  & 1.88 & \textbf{0.035} & \textbf{0.49°}& 0.896& 2640& 95.6\\
       & Ours & \textbf{0.93} & 0.060& 2.34°& \textbf{0.897}& \textbf{14.7}& \textbf{98.5}\\
\bottomrule
\end{tabular}
}
\end{table}

\subsection{Multi-Instance Matching Performance}

To further evaluate the generalization capability of our method, we construct a set of test samples containing multiple template instances per image using a controlled synthetic strategy. Specifically, we paste template objects into randomly selected regions of the search image to form multi-instance test samples. It is important to note that our model is trained solely on single-instance data and directly applied to multi-instance settings without any fine-tuning, in order to assess its cross-scenario generalization ability.

Given that traditional NCC-based method and Fast-Match have already shown significant disadvantages in both accuracy and efficiency during single-instance experiments, the following comparisons focus exclusively on Halcon’s SHM to evaluate deployment-level competitiveness, omitting NCC and Fast-Match for brevity. During sample generation, Each $320 \times 320$ search image is divided into a $4 \times 4$ grid (16 regions in total). Templates are cropped from separate source images and pasted into random grid locations, with random in-plane rotation applied within a range of $\pm180^\circ$. The pasted targets are scaled to a standard template size (36$\times$36) based on their original dimensions, and the number of pasted instances is randomized per image. This synthesis strategy simulate practical multi-instance inspection scenarios where multiple identical components may appear simultaneously with varying poses.

\begin{figure}[H]
\centering
\includegraphics[width=1\linewidth]{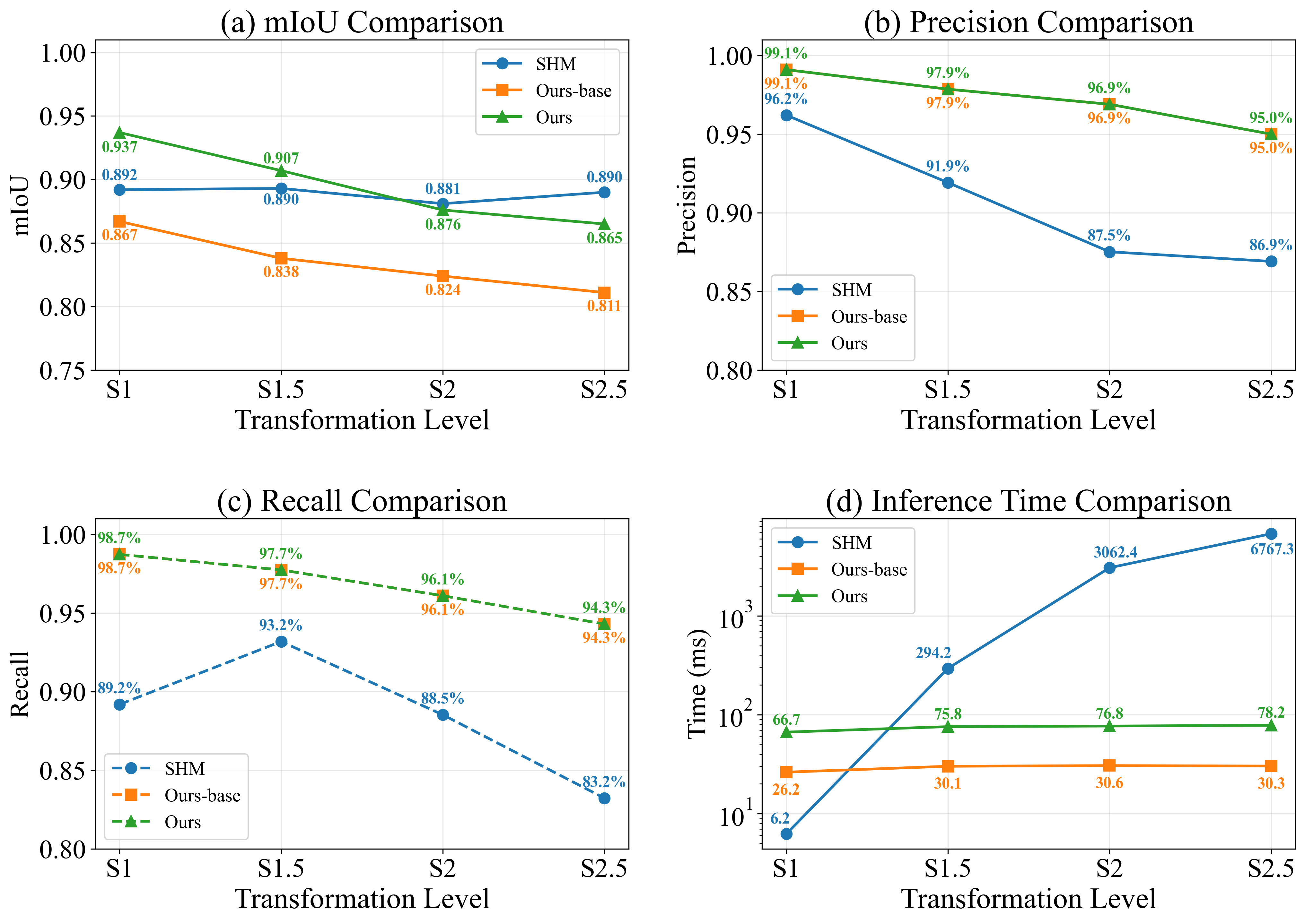}
\vspace{-2mm}
\caption{Performance comparison in multi-target matching scenarios. Our method achieves the highest precision and recall across all transformation levels and surpasses SHM in precision and mIoU under mild and moderate transformations (S1–S1.5). It also offers better inference efficiency in compound transformation scenarios. }
\label{fig:multi}
\vspace{-3mm}
\end{figure}

As shown in Figure~\ref{fig:multi}, our method exhibits strong robustness and inference efficiency across all transformation levels in the multi-instance setting. While SHM maintains stable accuracy as transformation intensity increases, its reliance on exhaustive parameter enumeration causes inference time to grow rapidly. Under the S2.5 transformation level, its latency becomes impractical for real-time industrial applications. In contrast, our approach maintains consistent computational cost across all levels thanks to the end-to-end geometric regression framework. Although mIoU slightly lags behind SHM under S2-S2.5 conditions, our method consistently achieves higher precision and recall across all settings, indicating superior stability and detection consistency in multi-target scenarios. Additionally, ours-base provides direct regression head outputs (rotation and scale parameters) without refinement module processing. Overall, these results demonstrate that our method offers a more favorable balance of speed, robustness, and deployment flexibility, particularly for industrial tasks involving multiple instances with complex pose variations.

\subsection{Adaptability to Varying Template Sizes}

To assess the model’s adaptability to different template sizes, we conduct experiments under various template configurations. Specifically, we evaluate the proposed method using four representative template sizes: $12 \times 12$, $28 \times 52$, $36 \times 36$, and $124 \times 124$. All experiments in this section are performed under the S1.5 transformation level, with other training configurations kept unchanged to ensure fair comparison.

As shown in Table~\ref{tab:template_sizes}. Experimental results indicate that the proposed method exhibits strong adaptability across all template sizes. In particular, under the small-sized templates setting ($12 \times 12$), our method maintains high matching accuracy and reliable detection capability. In contrast, SHM exhibits a high failure rate when matching small targets, with the success rate dropping to only 76.6\%, and significantly larger errors in both rotation and scale estimation. Benefiting from the template-aware modeling and refinement mechanism, our method achieves more accurate position prediction under such challenging conditions. Under large-template settings ($124 \times 124$), SHM outperforms our method in both accuracy and inference speed, owing to its edge-based shape modeling approach. With high-resolution boundary features available, SHM’s matching mechanism becomes more effective and efficient. In comparison, our method experiences a slight drop in precision and a notable increase in computation time due to the higher cost of dynamic convolution and refinement when processing larger templates. This highlights that SHM still holds advantages when the template quality is high and boundary features are rich. For medium-sized templates ($28 \times 52$ and $36 \times 36$), our method achieves significantly lower localization error and higher mIoU compared to SHM, with a \textbf{5–7× }speed advantage and comparable success rates. Overall, the proposed method excels in small-object scenarios, delivering more precise localization and enhanced robustness. These advantages make it particularly well-suited for real-world applications such as high-density layouts and small-object industrial inspection.

\begin{table}[htbp]
\centering
\caption{Performance under different template sizes.}
\label{tab:template_sizes}
\renewcommand{\arraystretch}{1.3} 
\resizebox{\textwidth}{!}{
\begin{tabular}{c l c c c c c c}
\toprule
\textbf{Size} & \textbf{Method} & \textbf{Loc.Err}↓& \textbf{ScaleErr}↓& \textbf{Rot.Err}↓& \textbf{mIoU}↑ & \textbf{Time(ms)}↓& \textbf{Succ.(\%)}↑\\
\midrule
12×12   & SHM   & 1.62 & \textbf{0.123} & \textbf{5.24°}& 0.722& 27.8& 76.6\\
        & Ours & \textbf{0.70}& 0.135& 5.89°& \textbf{0.808}& \textbf{12.2}& \textbf{97.3}\\
28×52   & SHM   & 1.65 & \textbf{0.015} & \textbf{0.21°}& 0.907& 67.5& 99.2\\
        & Ours & \textbf{0.55} & 0.036& 1.04°& \textbf{0.934}& \textbf{13.5}& \textbf{99.3}\\
36×36   & SHM   & 1.50 & \textbf{0.027} & \textbf{0.47°}& 0.908& 97.8& 97.8\\
        & Ours & \textbf{0.64}& 0.038& 1.63°& \textbf{0.926}& \textbf{13.8}& \textbf{98.6}\\
124×124 & SHM   & 1.56& \textbf{0.004} & \textbf{0.04°}& \textbf{0.969}& \textbf{26.3}& 98.6\\
        & Ours & \textbf{1.55}& 0.041& 1.05°& 0.945& 38.4& \textbf{98.9}\\
\bottomrule
\end{tabular}
}
\end{table}

\subsection{Evaluation on Real-World Industrial Data}
\label{sec:Real-World Industrial Data}
To evaluate the deployability of the proposed method in practical industrial applications, we conducted direct testing on a real-world defect inspection task. In this scenario, each image contains 144 repeated targets, and the image resolution is adjusted such that each target has an approximate size of $28 \times 52$ pixels. This dataset involves only in-plane rotation without scale variation, and the background is relatively clean and free of clutter. The inspected objects have fine but repetitive textures, and the images were captured under stable illumination with minimal lighting variation, making the evaluation environment relatively controlled. It is important to emphasize that we did not perform any additional training or hyperparameter tuning. The inference was carried out directly using the same general-purpose model trained on synthetic data. Experimental results show that the proposed method achieves 99\% in both precision and recall on this dataset, with an average IoU of 0.95 and an inference time of 2.28 seconds per image. These results demonstrate that the model delivers high accuracy and stability in structured, low-noise environments, fulfilling the accuracy and latency requirements of industrial online inspection systems. Note that the industrial dataset used in this evaluation is proprietary and cannot be publicly released due to confidentiality agreements. As such, we report only performance metrics without disclosing sample images. To mitigate confidentiality constraints that preclude releasing industrial data, future work could explore releasing synthetic industrial-like benchmarks or providing anonymized image patches that preserve task-relevant geometry and texture while removing sensitive information. 

\subsection{Evaluation on DOTA-v1 Dataset for Cross-Domain Generalization}
To further evaluate the cross-domain generalization capability of the proposed method, we conduct additional experiments on the DOTA-v1 dataset, which contains high-resolution aerial images with rotated bounding box annotations across diverse backgrounds and object scales. For each annotated instance, the rotated bounding box is first rectified to a horizontal orientation to obtain a canonical template, which is then matched back onto the original image using our proposed framework.

Since many large DOTA images contain multiple visually similar objects within the same scene, directly performing full-image matching may lead to unreasonable mismatches. To mitigate this issue, we restrict the search region to a $224\times224$ crop centered on the annotated instance, which effectively reduces distractors and normalizes the effective scale range for matching. Importantly, our model is neither trained nor fine-tuned on DOTA-v1, ensuring a strict cross-domain evaluation setting 

We compare our method against SHM, while we omit NCC and Fast-Match here because their performance on DOTA-v1 is significantly lower and not competitive in high-resolution aerial imagery. As shown in Table~\ref{tab:dota_results}, our method achieves consistently lower localization errors than SHM under all settings, with errors remaining below 1 pixel, and delivers substantially faster inference on S1.5 to S2.5, while maintaining comparable or superior matching success rates.

\begin{table}[H]
\centering
\caption{Performance on the DOTA-v1 dataset under different transformation settings.}
\label{tab:dota_results}
\renewcommand{\arraystretch}{1.3} 
\resizebox{\textwidth}{!}{
\begin{tabular}{c l c c c c c c}
\toprule
\textbf{Setting} & \textbf{Method} & \textbf{Loc.Err}↓ & \textbf{ScaleErr}↓ & \textbf{Rot.Err}↓ & \textbf{mIoU}↑ & \textbf{Time(ms)}↓ & \textbf{Succ.(\%)}↑ \\
\midrule
S1   & SHM              & 1.61& --& \textbf{0.30°} & 0.880& \textbf{6.87} & 96.3 \\
     & Ours& \textbf{0.62}& --& 1.61° & \textbf{0.911}& 11.2 & \textbf{97.8} \\
S1.5 & SHM              & 1.75& \textbf{0.043}& \textbf{0.36°} & 0.869& 108  & 95.8 \\
     & Ours& \textbf{0.67}& 0.052& 2.95°& \textbf{0.894}& \textbf{13.9}& \textbf{97.9} \\
S2   & SHM              & 1.95 & \textbf{0.041}& \textbf{0.33°} & \textbf{0.864}& 922  & 95.2\\
     & Ours& \textbf{0.92}& 0.047& 5.28° & 0.843& \textbf{14.6} & \textbf{96.1}\\
S2.5 & SHM              & 2.02& \textbf{0.046}& \textbf{0.43°} & \textbf{0.862}& 2485 & \textbf{94.5} \\
     & Ours& \textbf{0.98}& 0.056& 5.58° & 0.839& \textbf{15.4} & 94.2 \\
\bottomrule
\end{tabular}
}
\end{table}

\subsection{Case Study: Analysis of Typical Failure Cases}

\begin{figure}[H]
\centering
\includegraphics[width=1\linewidth]{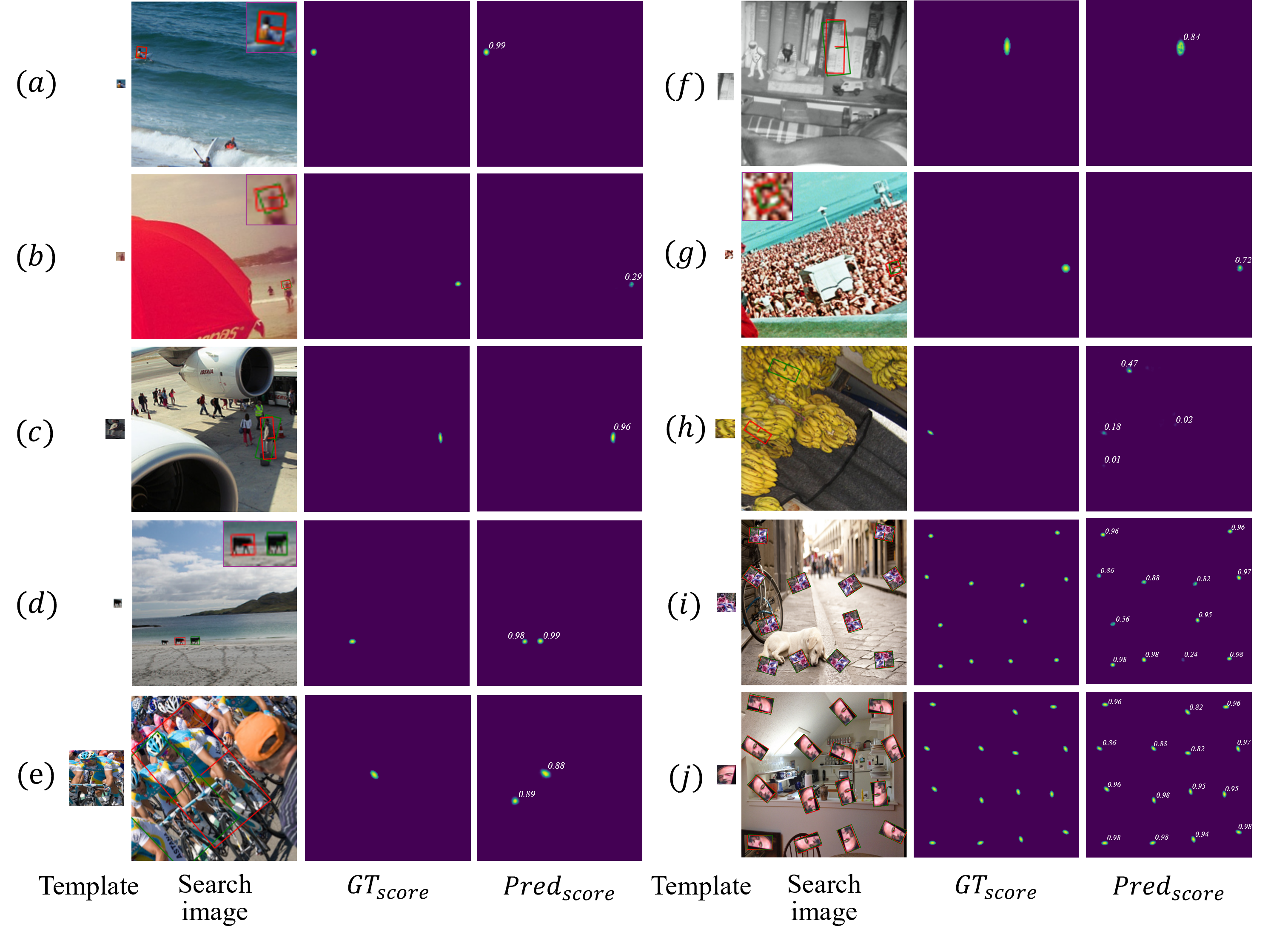}
\vspace{-2mm}
\caption{Typical failure and challenging cases. Each subfigure shows the input template, the corresponding search image, and the predicted results. The red bounding box denotes the ground-truth position, while the green bounding box indicates the predicted location. \textbf{ \(GT_{score}\)} refers to the ground-truth center heatmap generated via affine transformation. \textbf{ \(pred_{score}\)} denotes predicted center heatmap. (a) normal; (b) Tiny template; (c) Severe aspect ratio distortion; (d–e) Highly similar distractors; (f) low-texture template; (g-h) background clutter with similar patterns; (i-j) multi-object suppression. }
\label{fig:case_study}
\vspace{-3mm}
\end{figure}

To further investigate the limitations of the proposed method in real-world scenarios, we conduct a qualitative case study on representative failure instances, as shown in Figure~\ref{fig:case_study}. Each case illustrates, from left to right: the input template and search image (with the ground-truth location marked in red and the predicted location in green), the matching confidence within the ground-truth region ($\text{GT}_{\text{score}}$), and the predicted high-response region in the score map ($\text{pred}_{\text{score}}$). It should be noted that these visualizations focus solely on center localization heatmaps and do not include the regressed geometric parameters. The observed failure cases can be broadly categorized into five representative types:

\begin{itemize}
  \item \textbf{Low-texture templates:} Lack of distinctive texture leads to unstable responses and reduced precision.
  
\item \textbf{Tiny templates:} Extremely small target regions hinder accurate detection and lower confidence. 
  
  \item \textbf{Severe aspect ratio distortions:} Extreme shape changes disrupt feature alignment and reduce regression accuracy.
  
  \item \textbf{Background clutter with similar patterns:} Repetitive textures cause confusion, lowering confidence and target discrimination. 
  
  \item \textbf{Highly similar distractors:} Look-alike regions cause false positives and ambiguous localization.

  \item \textbf{Multi-object suppression}: Missing multi-target supervision leads to suppressed responses and lower accuracy.
\end{itemize}

Despite these challenging scenarios, the proposed method performs robustly in most cases with sufficient texture, clear boundaries, and moderate geometric variations. These results highlight potential areas for further enhancement, such as incorporating saliency-aware attention, multi-template fusion, or context-based refinement to improve matching robustness under adverse conditions.

\subsection{Ablation Study}

To provide a comprehensive understanding of our framework, this section evaluates the contributions of individual components and investigates the trade-offs between model capacity and runtime efficiency. For consistency, unless otherwise specified, all experiments in this section are conducted using $36\times36$ templates under the S1.5 transformation setting on the MS-COCO dataset.  For clarity, this section is organized into two parts, with the first focusing on Component Ablations and the second on Model Capacity and Efficiency Analysis.

\subsubsection{Component Ablations}
\paragraph{Effect of Geometric Refinement}
We first assess the contribution of the geometric refinement module. As shown in Table~\ref{tab:ablation_refine_heatmap}, the model without refinement (Ours-base) directly outputs the predicted rotation and scale parameters from the regression head. This baseline achieves decent localization but suffers from significant geometric estimation errors. Incorporating the refinement module improves both rotation and scale accuracy, with mIoU increasing from 0.905 to 0.926 and both rotation and scale errors decreasing by approximately 60\%. It is worth noting that this refinement module is optional: when applied on backbone predictions, it introduces an additional computational overhead of approximately 2.7\,ms per image pair (around \textbf{25\%} of runtime), which remains acceptable given the observed accuracy gains.

\begin{table}[htbp]
\centering
\caption{Ablation study on the refinement module and heatmap supervision.}
\label{tab:ablation_refine_heatmap}
\resizebox{\textwidth}{!}{%
\begin{tabular}{l|ccccccc}
\toprule
\textbf{Variant} & \textbf{Heatmap Type} & \textbf{Refine} & \textbf{Loc.Err}↓&  \textbf{ScaleErr}↓
&\textbf{Rot.Err}↓& \textbf{mIoU}↑ & \textbf{Time(ms)}↓\\
\midrule
Isotropic          & Isotropic  & \ding{55}     & 0.66&  0.109
&5.12°& 0.899& \textbf{11.1}\\
Ours-base& Elliptical & \ding{55}     & \textbf{0.64}&  0.083
&4.38°& 0.905& \textbf{11.1}\\
Ours (full)        & Elliptical & \ding{51} & \textbf{0.64}&  \textbf{0.038}&\textbf{1.93°}& \textbf{0.926}& 13.8\\
\bottomrule
\end{tabular}%
}
\end{table}

\begin{figure}[H]   
\centering
\includegraphics[width=\linewidth]{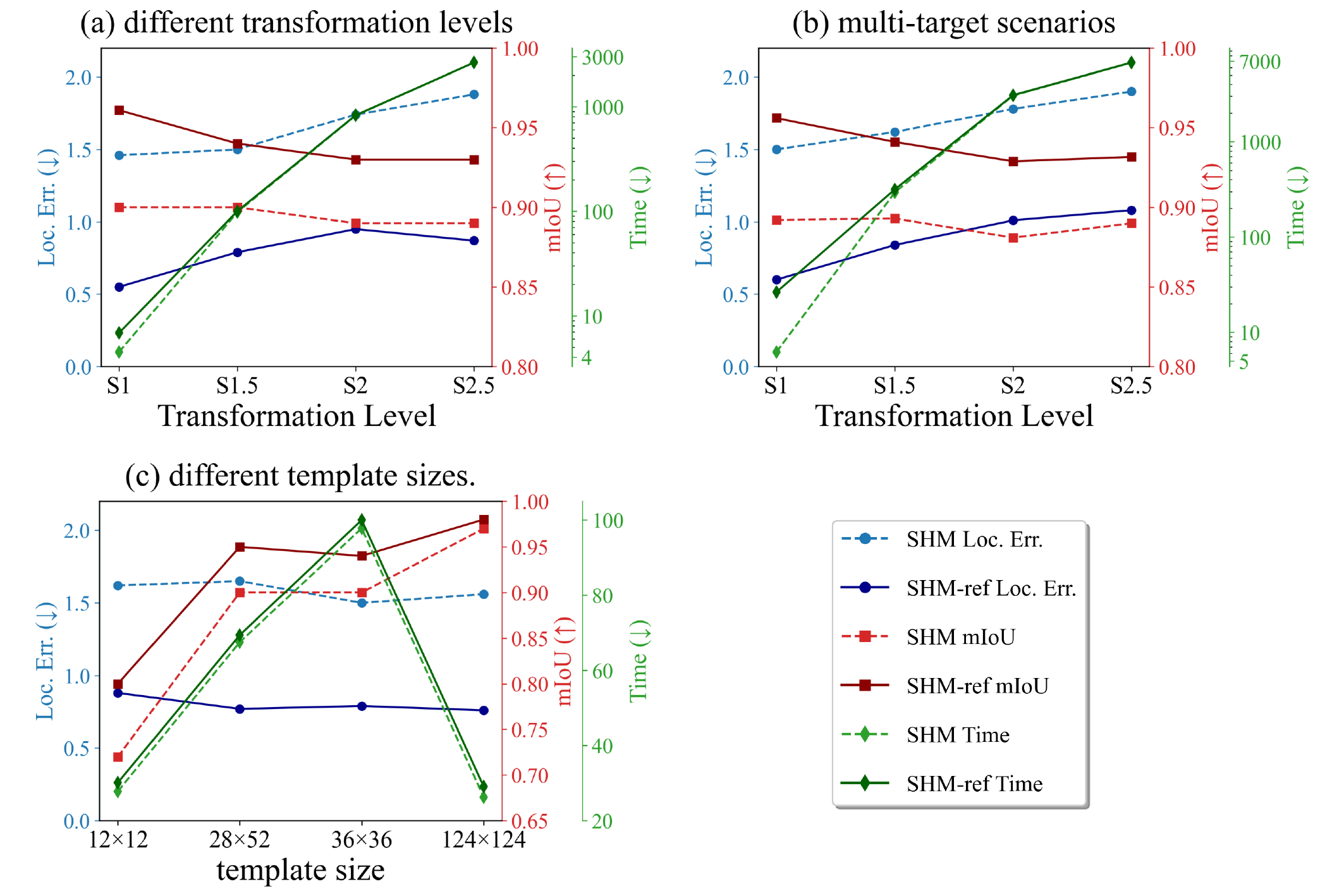}
\caption{Evaluation of the proposed refinement module when applied to SHM results. The plots show performance comparisons across (a) different transformation levels, (b) multi-instance scenarios, and (c) varying template sizes. In all cases, the refined version (-ref) achieves lower localization error and higher mIoU.}
\label{fig:refineShm}
\end{figure}

To demonstrate the generality of our geometric refinement module, we apply it directly on the initial predictions produced by SHM. Since SHM already achieves high precision in estimating scale and rotation, but exhibits relatively large localization error, we perform refinement exclusively on the predicted position while keeping other geometric parameters unchanged. Figure~\ref{fig:refineShm} presents three representative scenarios where the SHM predictions serve as coarse estimates, and our refinement module is applied to improve localization quality: (a) different transformation levels,  (b) multi-instance scenarios,  (c) Varying template sizes. Each case visualizes the effect of refinement on three key indicators: localization error, mIoU, and inference time. Across these scenarios:

\begin{itemize}
    \item Localization error is reduced by 40--60\textbf{\%}, e.g., from 1.5 px to 0.6 px under mild transformation;
    \item mIoU improves by 4--6\%, e.g., from 0.89 to 0.94;
    \item In most cases, the inference time increases by only 1--2\% relative to the original runtime.
\end{itemize}
These results demonstrate that our refinement module significantly improves geometric alignment across diverse conditions, with negligible computational overhead. The consistent gains confirm its broad applicability as a lightweight and general-purpose post-processing tool.

\paragraph{Geometry-Aware Heatmap Supervision}
To validate the effectiveness of the elliptical heatmap design, we replace it with an isotropic Gaussian, where the supervision only encodes center location without orientation or aspect ratio. As shown in Table~\ref{tab:ablation_refine_heatmap}, removing the geometric encoding degrades both the regression accuracy and spatial alignment, confirming the importance of structure-aware supervision in training the network.

\subsubsection{Model Capacity and Efficiency Analysis}
\paragraph{Effect of Feature Depth}
To assess the effect of feature abstraction on both accuracy and runtime, we compare two backbone configurations: a shallow feature extractor (ConvNeXt-V2-Tiny~\cite{convNextV2_2023}  Stage-1, used in our default design) and a deeper alternative (Stage-2). Both variants use the same network head and training strategy. As shown in Table~\ref{tab:backbone_depth}, Stage-2 offers marginal gains in localization and pose accuracy over Stage-1, but with more than twice the inference latency (34.7 ms vs. 13.8 ms). This demonstrates that our Stage-1 configuration strikes a better balance between accuracy and speed, making it more suitable for time-sensitive applications. 

\begin{table}[H]
\centering
\caption{Impact of backbone feature depth. }
\label{tab:backbone_depth}
\resizebox{\textwidth}{!}{  
\begin{tabular}{l|cccccc}
\toprule
\textbf{Feature Stage} & \textbf{Loc.Err} ↓ &  \textbf{ScaleErr} ↓ 
&\textbf{Rot.Err} ↓& \textbf{mIoU} ↑ & \textbf{Time(ms)} ↓& \textbf{Params(M)} ↓\\
\midrule
Stage-1 (Ours)   & 0.64&  0.038
&1.63°& 0.926& \textbf{13.8}& \textbf{3.07}\\
Stage-2 (Deep)   & \textbf{0.61}&  \textbf{0.033}&\textbf{1.59°}& \textbf{0.928}& 34.7& 14.21\\
\bottomrule
\end{tabular}
}
\end{table}

\paragraph{Effect of Backbone Architecture}
We compare four ConvNeXt-V2 backbones (Tiny, Base, Large, Huge) to evaluate the impact of model capacity on matching performance. As shown in Table~\ref{tab:backbone_type}, Tiny has slightly higher rotation and scale errors than larger variants but achieves comparable localization precision with the fastest inference and smallest model size. This trade-off makes it well-suited for real-time industrial applications, and thus we adopt ConvNeXt-V2-Tiny as the default backbone in our framework. 

\begin{table}[H]
\centering
\caption{Performance of different ConvNeXt-V2 backbones.}
\label{tab:backbone_type}
\resizebox{\textwidth}{!}{
\begin{tabular}{l|ccccccc}
\toprule
\textbf{Backbone} &   \textbf{Loc.Err}↓& \textbf{ScaleErr}↓
&\textbf{Rot.Err}↓& \textbf{mIoU}↑& \textbf{Succ.(\%)}↑& \textbf{Time(ms)}↓& \textbf{Params}↓\\
\midrule
Tiny(Ours)&   0.64& 0.038
&1.93°& 0.926& 99.0& \textbf{13.8}& \textbf{3.07M}\\
Base&   0.65& 0.037
&1.72°& 0.930 & \textbf{99.2}& 19.5& 5.35M\\
Large&   0.63& 0.038
&1.63°& 0.936& 98.6& 33.2& 11.85M\\
Huge&   \textbf{0.62}& \textbf{0.034}&\textbf{1.59°}& \textbf{0.939}& 98.4& 102& 41.92M\\
\bottomrule
\end{tabular}
}
\end{table}
\paragraph{Sensitivity Analysis of Refinement Hyperparameters}
\label{sec:sensitivity_refine}

To further investigate the influence of hyperparameters in the geometric refinement module, we conduct a sensitivity analysis on two key factors: the step size and the search range. These two parameters control the sampling resolution and search scope during refinement, which jointly affect both accuracy and runtime efficiency.

\begin{figure}[H]
    \centering
    \includegraphics[width=0.95\linewidth]{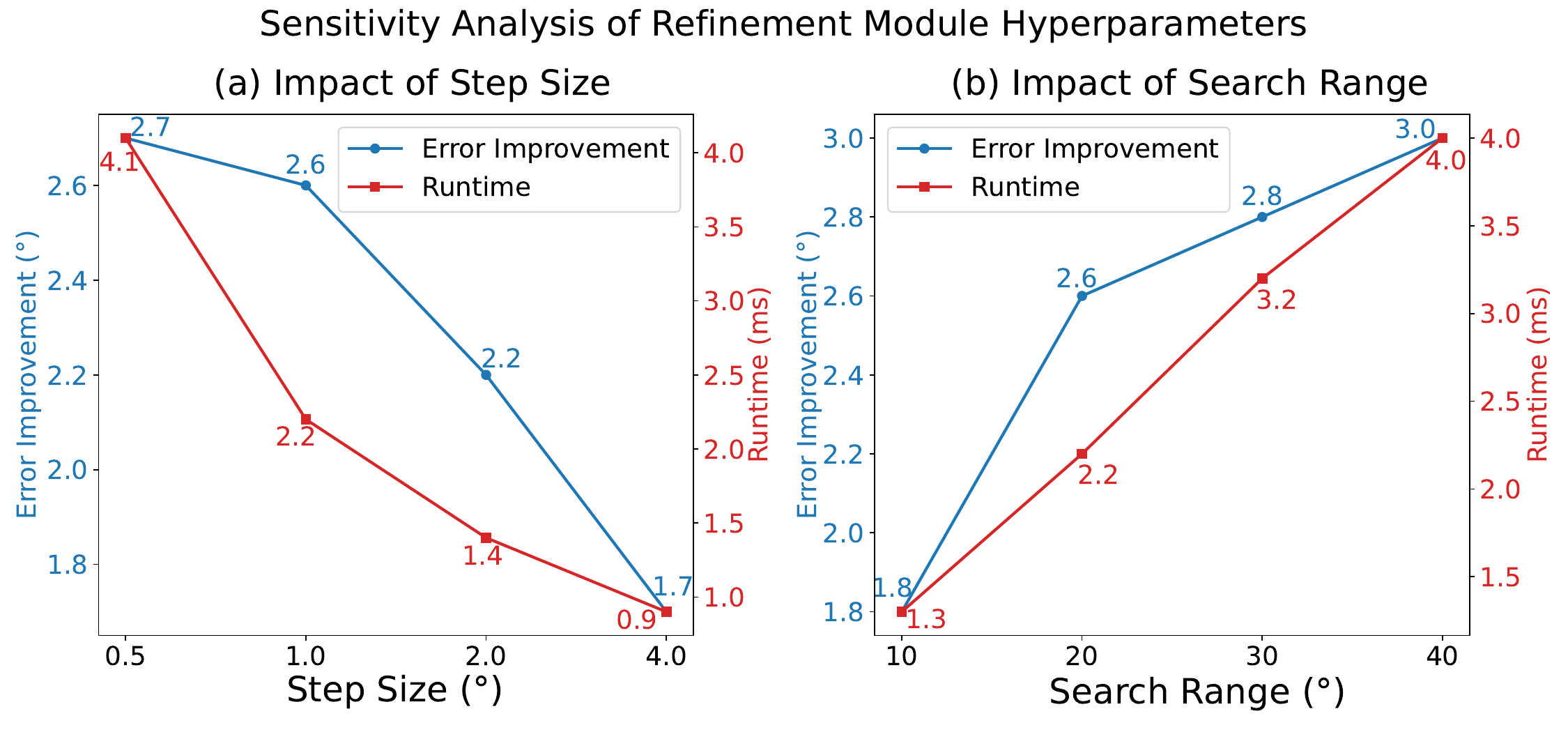}
    \caption{
    Sensitivity analysis of the geometric refinement module hyperparameters.
    \textbf{(a)} Impact of step size on error improvement and runtime.
    \textbf{(b)} Impact of search range on error improvement and runtime.
    ``Error improvement'' refers to the reduction in angle error compared to the raw backbone outputs.
    }
    \label{fig:sensitivity}
\end{figure}

For the step size analysis, we fix the search range to $\pm 20^\circ$ and vary the step size within $\{0.5^\circ, 1^\circ, 2^\circ, 4^\circ\}$. We measure two metrics: (\textit{i}) the error improvement, defined as the reduction in angle error compared with the raw backbone predictions, and (\textit{ii}) the runtime per image pair. As shown in Fig.~\ref{fig:sensitivity}(a), smaller step sizes yield higher error improvements but require longer computation times, whereas larger step sizes reduce runtime but slightly degrade accuracy.

For the search range analysis, we fix the step size to $1^\circ$ and vary the search range within $\{\pm 10^\circ, \pm 20^\circ, \pm 30^\circ\, \pm 40^\circ\}$. As shown in Fig.~\ref{fig:sensitivity}(b), wider search ranges provide larger accuracy gains but also lead to increased computational costs. Based on these results, we choose step=$1^\circ$ and range=$\pm 20^\circ$ as the default configuration, achieving a good balance between performance and efficiency.

\section{Conclusion}

This paper presents a lightweight, end-to-end deep template matching method that enables unified regression of both spatial location and geometric transformation parameters of the target. The core of the approach lies in the proposed Template-Aware Dynamic Convolution Module (TDCM), which models the template as a dynamically generated convolutional kernel to guide geometric response generation in the search image. In addition, a lightweight prediction structure is built using shallow features, depthwise separable convolutions, and pixel shuffle mechanisms, which significantly improves inference efficiency while maintaining high matching accuracy. A self-supervised training framework is further introduced based on rotation-shear transformations and spatial constraints, enabling automatic generation of high-quality pseudo-labels under geometric-annotation-free conditions. To improve the reliability of geometric predictions, a post-prediction refinement module is incorporated to locally optimize rotation and scale estimates.

Experimental results demonstrate that under compound geometric transformations involving both rotation and scaling, the proposed method achieves a 7–180× inference speed-up compared to SHM. In multi-instance matching tasks, the method exhibits strong target separation and consistent geometric modeling. The adaptability tests across different template sizes further verify its robustness and stability under small-sized templates conditions. In summary, the proposed approach suits industrial template matching under wide-scale variation, high-frequency, and complex deformations, offering real-time inference and strong deployment potential.

However, there are still limitations that warrant further investigation. The model is trained within a predefined template size range and may require fine-tuning when encountering templates whose scales fall outside the training distribution. This constraint limits its generalization across arbitrarily sized templates without adaptation. Furthermore, the model remains sensitive to non-rigid variations such as illumination changes, occlusions, and perspective distortions, which may degrade the matching performance in uncontrolled environments. Future work will explore several directions to overcome these limitations. First, we plan to develop a scale-adaptive architecture that generalizes across a broader range of template sizes without retraining. Second, we aim to enhance robustness against non-rigid and photometric deformations by incorporating deformation-invariant feature learning and domain generalization strategies.

\section*{Acknowledgments}

This work was supported by the Sichuan Science and Technology Support Program under Grant No. 2024YFG0001. 


\begin{thebibliography}{10}
\expandafter\ifx\csname url\endcsname\relax
  \def\url#1{\texttt{#1}}\fi
\expandafter\ifx\csname urlprefix\endcsname\relax\def\urlprefix{URL }\fi
\expandafter\ifx\csname href\endcsname\relax
  \def\href#1#2{#2} \def\path#1{#1}\fi

\bibitem{Barone2020}
M.~Barone, Image wafer inspection based on template matching, in: 10th International Conference on Digital Image Processing and Pattern Recognition DPPR, in press, 2020, pp. 45--56.
\newblock \href {https://doi.org/10.5121/csit.2020.101505} {\path{doi:10.5121/csit.2020.101505}}.

\bibitem{Chen2025}
Z.~Chen, X.~Xie, L.~Yang, J.-H. Lai, Hard-normal example-aware template mutual matching for industrial anomaly detection, International Journal of Computer Vision 133~(5) (2025) 2927--2949.
\newblock \href {https://doi.org/10.1007/s11263-024-02323-0} {\path{doi:10.1007/s11263-024-02323-0}}.

\bibitem{BrakeYang2025}
Y.~Yang, J.~Guo, C.~Liu, H.~Qian, P.~Gu, Brake disc defect detection method based on an improved multi-scale template matching algorithm, Signal, Image and Video Processing 19~(6) (2025) 484.
\newblock \href {https://doi.org/10.1007/s11760-025-04093-9} {\path{doi:10.1007/s11760-025-04093-9}}.

\bibitem{Le2020APCBAlignmentSystem}
M.-T. Le, C.-T. Tu, S.-M. Guo, J.-J.~J. Lien, A pcb alignment system using rst template matching with cuda on embedded gpu board, Sensors 20~(9) (2020).
\newblock \href {https://doi.org/10.3390/s20092736} {\path{doi:10.3390/s20092736}}.

\bibitem{Shahsavarani2024}
S.~Shahsavarani, F.~Lopez, C.~Ibarra-Castanedo, X.~P.~V. Maldague, Robust multi-modal image registration for image fusion enhancement in infrastructure inspection, Sensors 24~(12) (2024).
\newblock \href {https://doi.org/10.3390/s24123994} {\path{doi:10.3390/s24123994}}.

\bibitem{SSD2020}
W.~Sidi, I.~G. Wibawa, Sum of squared difference (ssd) template matching testing on writing learning application, JELIKU (Jurnal Elektronik Ilmu Komputer Udayana) 8 (2020) 453.
\newblock \href {https://doi.org/10.24843/JLK.2020.v08.i04.p11} {\path{doi:10.24843/JLK.2020.v08.i04.p11}}.

\bibitem{SSD-NCC}
M.~Hisham, S.~N. Yaakob, R.~Raof, A.~A. Nazren, N.~Wafi, Template matching using sum of squared difference and normalized cross correlation, in: 2015 IEEE Student Conference on Research and Development (SCOReD), 2015, pp. 100--104.
\newblock \href {https://doi.org/10.1109/SCORED.2015.7449303} {\path{doi:10.1109/SCORED.2015.7449303}}.

\bibitem{NCC}
D.-M. Tsai, C.-T. Lin, J.-F. Chen, The evaluation of normalized cross correlations for defect detection, Pattern Recognition Letters 24~(15) (2003) 2525--2535.
\newblock \href {https://doi.org/https://doi.org/10.1016/S0167-8655(03)00098-9} {\path{doi:https://doi.org/10.1016/S0167-8655(03)00098-9}}.

\bibitem{RLTM2007}
K.~Fredriksson, V.~M{\"a}kinen, G.~Navarro, Rotation and lighting invariant template matching, Information and Computation 205~(7) (2007) 1096--1113.
\newblock \href {https://doi.org/10.1016/j.ic.2007.03.002} {\path{doi:10.1016/j.ic.2007.03.002}}.

\bibitem{le2022robot}
M.-T. Le, J.-J.~J. Lien, Robot arm grasping using learning-based template matching and self-rotation learning network, The International Journal of Advanced Manufacturing Technology 121~(3) (2022) 1915--1926.
\newblock \href {https://doi.org/10.1007/s00170-022-09374-y} {\path{doi:10.1007/s00170-022-09374-y}}.

\bibitem{ttm2024}
A.~Martinez-Sanchez, U.~Homberg, J.~M. Almira, H.~Phelippeau, Tensorial template matching for fast cross-correlation with rotations and its application for tomography, in: Computer Vision -- ECCV 2024, Springer Nature Switzerland, 2024, pp. 19--35.
\newblock \href {https://doi.org/10.1007/978-3-031-73383-3_2} {\path{doi:10.1007/978-3-031-73383-3_2}}.

\bibitem{FNCCR2024}
J.~M. Almira, H.~Phelippeau, A.~Martinez-Sanchez, Fast normalized cross-correlation for template matching with rotations, Journal of Applied Mathematics and Computing 70~(5) (2024) 4937--4969.
\newblock \href {https://doi.org/10.1007/s12190-024-02157-6} {\path{doi:10.1007/s12190-024-02157-6}}.

\bibitem{FMT2005}
X.~Guo, Z.~Xu, Y.~Lu, Y.~Pang, An application of fourier-mellin transform in image registration, in: The Fifth International Conference on Computer and Information Technology (CIT'05), 2005, pp. 619--623.
\newblock \href {https://doi.org/10.1109/CIT.2005.62} {\path{doi:10.1109/CIT.2005.62}}.

\bibitem{GrayScale2007}
H.~Y. Kim, S.~A. de~Ara{\'u}jo, Grayscale template-matching invariant to rotation, scale, translation, brightness and contrast, in: D.~Mery, L.~Rueda (Eds.), Advances in Image and Video Technology, Springer Berlin Heidelberg, Berlin, Heidelberg, 2007, pp. 100--113.

\bibitem{TMbaseOrientGradient2012}
Y.~Konishi, Y.~Kotake, Y.~Ijiri, M.~Kawade, Fast and precise template matching based on oriented gradients, in: A.~Fusiello, V.~Murino, R.~Cucchiara (Eds.), Computer Vision -- ECCV 2012. Workshops and Demonstrations, Springer Berlin Heidelberg, Berlin, Heidelberg, 2012, pp. 607--610.

\bibitem{CHEN2016207}
C.-S. Chen, C.-L. Huang, C.-W. Yeh, W.-C. Chang, An accelerating cpu based correlation-based image alignment for real-time automatic optical inspection, Computers \& Electrical Engineering 49 (2016) 207--220.
\newblock \href {https://doi.org/https://doi.org/10.1016/j.compeleceng.2015.09.010} {\path{doi:https://doi.org/10.1016/j.compeleceng.2015.09.010}}.

\bibitem{FAstMatch2013}
S.~Korman, D.~Reichman, G.~Tsur, S.~Avidan, Fast-match: Fast affine template matching, in: 2013 IEEE Conference on Computer Vision and Pattern Recognition, 2013, pp. 2331--2338.
\newblock \href {https://doi.org/10.1109/CVPR.2013.302} {\path{doi:10.1109/CVPR.2013.302}}.

\bibitem{halcon2024}
{MVTec Software GmbH}, {\textit{HALCON} v24.11.1.0 [software]}, \url{https://www.mvtec.com/products/halcon/}, accessed: July 13, 2025 (2024).

\bibitem{qatm}
J.~Cheng, Y.~Wu, W.~AbdAlmageed, P.~Natarajan, Qatm: Quality-aware template matching for deep learning, in: 2019 IEEE/CVF Conference on Computer Vision and Pattern Recognition (CVPR), 2019, pp. 11545--11554.
\newblock \href {https://doi.org/10.1109/CVPR.2019.01182} {\path{doi:10.1109/CVPR.2019.01182}}.

\bibitem{SiameseSARwu2022}
W.~Wu, Y.~Xian, J.~Su, L.~Ren, A siamese template matching method for sar and optical image, IEEE Geoscience and Remote Sensing Letters 19 (2022) 1--5.
\newblock \href {https://doi.org/10.1109/LGRS.2021.3108579} {\path{doi:10.1109/LGRS.2021.3108579}}.

\bibitem{End-to-EndTMSiameseNetwork2022}
Q.~Ren, Y.~Zheng, P.~Sun, W.~Xu, D.~Zhu, D.~Yang, A robust and accurate end-to-end template matching method based on the siamese network, IEEE Geoscience and Remote Sensing Letters 19 (2022) 1--5.
\newblock \href {https://doi.org/10.1109/LGRS.2021.3094046} {\path{doi:10.1109/LGRS.2021.3094046}}.

\bibitem{CenterPointRemoteImagery2024}
J.~Yang, Y.~Zheng, W.~Xu, P.~Sun, S.~Bai, An accurate and robust multimodal template matching method based on center-point localization in remote sensing imagery, Remote Sensing 16~(15) (2024) 2831.
\newblock \href {https://doi.org/10.3390/rs16152831} {\path{doi:10.3390/rs16152831}}.

\bibitem{selfTM}
A.~Hristov, D.~Dimov, M.~Nisheva-Pavlova, Self-supervised foundation model for template matching, Big Data and Cognitive Computing 9~(2) (2025) 38.
\newblock \href {https://doi.org/10.3390/bdcc9020038} {\path{doi:10.3390/bdcc9020038}}.

\bibitem{SiameseNetworkYang2025}
J.~Yang, W.~Xu, Y.~Zheng, Q.~Ren, S.~Bai, An accurate template matching method based on siamese network and center point estimation, in: L.~Yan, H.~Duan, Y.~Deng (Eds.), Advances in Guidance, Navigation and Control, Springer Nature Singapore, Singapore, 2025, pp. 408--419.

\bibitem{sift2004}
D.~G. Lowe, Distinctive image features from scale-invariant keypoints, International Journal of Computer Vision 60~(2) (2004) 91--110.
\newblock \href {https://doi.org/10.1023/B:VISI.0000029664.99615.94} {\path{doi:10.1023/B:VISI.0000029664.99615.94}}.

\bibitem{rublee2011orb}
E.~Rublee, V.~Rabaud, K.~Konolige, G.~Bradski, Orb: An efficient alternative to sift or surf, in: 2011 International Conference on Computer Vision, Ieee, 2011, pp. 2564--2571.
\newblock \href {https://doi.org/10.1109/ICCV.2011.6126544} {\path{doi:10.1109/ICCV.2011.6126544}}.

\bibitem{superpoints2018}
D.~DeTone, T.~Malisiewicz, A.~Rabinovich, Superpoint: Self-supervised interest point detection and description, in: 2018 IEEE/CVF Conference on Computer Vision and Pattern Recognition Workshops (CVPRW), 2018, pp. 337--33712.
\newblock \href {https://doi.org/10.1109/CVPRW.2018.00060} {\path{doi:10.1109/CVPRW.2018.00060}}.

\bibitem{R2D22019}
J.~Revaud, P.~Weinzaepfel, C.~D. Souza, M.~Humenberger, R2d2: Repeatable and reliable detector and descriptor, in: Proceedings of the 33rd International Conference on Neural Information Processing Systems (NeurIPS), Curran Associates Inc., Red Hook, NY, USA, 2019, p. 1113.

\bibitem{BBS2015}
T.~Dekel, S.~Oron, M.~Rubinstein, S.~Avidan, W.~T. Freeman, Best-buddies similarity for robust template matching, in: 2015 IEEE Conference on Computer Vision and Pattern Recognition (CVPR), 2015, pp. 2021--2029.
\newblock \href {https://doi.org/10.1109/CVPR.2015.7298813} {\path{doi:10.1109/CVPR.2015.7298813}}.

\bibitem{DDS2017}
I.~Talmi, R.~Mechrez, L.~Zelnik-Manor, Template matching with deformable diversity similarity, 2017 IEEE Conference on Computer Vision and Pattern Recognition (CVPR) (2016) 1311--1319.

\bibitem{yoo2009fastCNN}
J.-C. Yoo, T.~H. Han, Fast normalized cross-correlation, Circuits, Systems and Signal Processing 28~(6) (2009) 819--843.
\newblock \href {https://doi.org/10.1007/s00034-009-9130-7} {\path{doi:10.1007/s00034-009-9130-7}}.

\bibitem{kaso2018fft_ncc}
A.~Kaso, Computation of the normalized cross-correlation by fast fourier transform, PLoS ONE 13~(9) (2018) e0203434.
\newblock \href {https://doi.org/10.1371/journal.pone.0203434} {\path{doi:10.1371/journal.pone.0203434}}.

\bibitem{diwu2018}
L.~Talker, Y.~Moses, I.~Shimshoni, Efficient sliding window computation for nn-based template matching, in: V.~Ferrari, M.~Hebert, C.~Sminchisescu, Y.~Weiss (Eds.), Computer Vision -- ECCV 2018, Springer International Publishing, Cham, 2018, pp. 409--424.
\newblock \href {https://doi.org/10.1007/978-3-030-01249-6_25} {\path{doi:10.1007/978-3-030-01249-6_25}}.

\bibitem{vqnnf2023}
A.~Gupta, I.-M. Sintorn, Efficient high-resolution template matching with vector quantized nearest neighbour fields, Pattern Recognit. 151 (2024) 110386.
\newblock \href {https://doi.org/10.1016/j.patcog.2024.110386} {\path{doi:10.1016/j.patcog.2024.110386}}.

\bibitem{segmentNCC}
D.~Marušić, S.~Popović, Z.~Kalafatić, Template matching in images using segmented normalized cross-correlation, arXiv preprint arXiv:2502.01286 (2025).
\newblock \href {https://doi.org/10.48550/ARXIV.2502.01286} {\path{doi:10.48550/ARXIV.2502.01286}}.

\bibitem{deepTM}
Z.~Gao, R.~Yi, Z.~Qin, Y.~Ye, C.~Zhu, K.~Xu, Learning accurate template matching with differentiable coarse-to-fine correspondence refinement, Computational Visual Media 10~(2) (2024) 309--330.
\newblock \href {https://doi.org/10.1007/s41095-023-0333-9} {\path{doi:10.1007/s41095-023-0333-9}}.

\bibitem{convNextV2_2023}
S.~Woo, S.~Debnath, R.~Hu, X.~Chen, Z.~Liu, I.~S. Kweon, S.~Xie, Convnext v2: Co-designing and scaling convnets with masked autoencoders, in: Proceedings of the IEEE/CVF Conference on Computer Vision and Pattern Recognition (CVPR), 2023, pp. 16133--16142.

\bibitem{CondConv2019}
B.~Yang, G.~Bender, Q.~V. Le, J.~Ngiam, Condconv: Conditionally parameterized convolutions for efficient inference, in: Advances in Neural Information Processing Systems, Vol.~32, Curran Associates, Inc., 2019.

\bibitem{DynamicFilterNet}
X.~Jia, B.~De~Brabandere, T.~Tuytelaars, L.~V. Gool, Dynamic filter networks, in: Advances in Neural Information Processing Systems, Vol.~29, Curran Associates, Inc., 2016.

\bibitem{DCNv4}
Y.~Xiong, Z.~Li, Y.~Chen, F.~Wang, X.~Zhu, J.~Luo, Efficient deformable convnets: Rethinking dynamic and sparse operator for vision applications, 2024, pp. 5652--5661.
\newblock \href {https://doi.org/10.1109/CVPR52733.2024.00540} {\path{doi:10.1109/CVPR52733.2024.00540}}.

\bibitem{simple2021cvpr}
Z.~Hu, Z.~Yang, X.~Hu, R.~Nevatia, Simple: Similar pseudo label exploitation for semi-supervised classification, 2021, pp. 15094--15103.
\newblock \href {https://doi.org/10.1109/CVPR46437.2021.01485} {\path{doi:10.1109/CVPR46437.2021.01485}}.

\bibitem{spatialpooling2022pr}
X.~Jin, Y.~Xie, X.-S. Wei, B.-R. Zhao, Z.-M. Chen, X.~Tan, Delving deep into spatial pooling for squeeze-and-excitation networks, Pattern Recognition 121 (2021) 108159.
\newblock \href {https://doi.org/10.1016/j.patcog.2021.108159} {\path{doi:10.1016/j.patcog.2021.108159}}.

\bibitem{fnnet2025pr}
S.~Ye, Q.~Peng, Y.-m. Cheung, Y.~Wang, Z.~Zou, X.~You, Fn-net: Adaptive data augmentation network for fine-grained visual categorization, Pattern Recognition (2025) 111618.

\end{thebibliography}

\end{document}